\documentclass[runningheads]{llncs}

\usepackage{eccv}

\usepackage{eccvabbrv}

\usepackage{graphicx}
\usepackage{booktabs}

\usepackage{tabularx}
\usepackage{siunitx} 
\usepackage{titletoc}
\usepackage{multirow}
\usepackage{adjustbox}
\usepackage[table]{xcolor} 
\usepackage{pifont} 
\usepackage{bm}
\usepackage{soul}
\usepackage{algorithm}
\usepackage{algpseudocode}
\definecolor{softpink}{RGB}{220, 65, 130}
\newcommand{\papername}{SignSparK}
\newcommand{\cmark}{\textcolor{green!60!black}{\ding{51}}} 
\newcommand{\xmark}{\textcolor{red!70!black}{\ding{55}}}

\usepackage[accsupp]{axessibility}  

\usepackage[pagebackref,breaklinks,colorlinks,citecolor=eccvblue]{hyperref}

\usepackage{orcidlink}

\usepackage{amsmath,amsfonts,bm}

\def\eqref#1{equation~\ref{#1}}

\def\1{\bm{1}}

\def\vu{{\bm{u}}}
\def\vv{{\bm{v}}}

\def\vx{{\bm{x}}}

\def\vz{{\bm{z}}}

\DeclareMathAlphabet{\mathsfit}{\encodingdefault}{\sfdefault}{m}{sl}
\SetMathAlphabet{\mathsfit}{bold}{\encodingdefault}{\sfdefault}{bx}{n}

\begin{document}
\title{SignSparK: Efficient Multilingual Sign Language Production via Sparse Keyframe Learning} 

\titlerunning{SignSparK}

\author{Jianhe Low\orcidlink{0009-0009-0452-4374} \and
Alexandre Symeonidis-Herzig\orcidlink{0009-0003-1688-5317} \and
Maksym Ivashechkin\orcidlink{0000-0003-4936-1344} \and
\\Ozge Mercanoglu Sincan\orcidlink{0000-0001-9131-0634} \and
Richard Bowden\orcidlink{0000-0003-3285-8020}}

\authorrunning{J.H.Low et al.}


\institute{CVSSP, University of Surrey, United Kingdom
\\ \small{\texttt{\{jianhe.low, a.symeonidisherzig, 
m.ivashechkin,\\o.mercanoglusincan, r.bowden\}@surrey.ac.uk}}\\
\url{https://cogvis-cvssp.github.io/papers/signspark/}
}

\maketitle

\begin{abstract}
    Sign Language Production (SLP) faces a fundamental trade-off: direct text-to-pose models suffer from regression-to-the-mean effects, while dictionary-retrieval methods produce disjointed transitions. To resolve this, we propose a novel training paradigm that leverages sparse keyframes to capture the underlying kinematic distribution of human signing. By generating dense motion from discrete anchors, our approach mitigates regression-to-the-mean while ensuring fluid articulation. To achieve this at scale, we introduce FAST, an ultra-efficient sign segmentation model that automatically mines precise temporal boundaries. We then present \papername{}, a Conditional Flow Matching (CFM) framework that utilizes these temporal anchors to synthesize 3D signing sequences. This keyframe-driven formulation also unlocks Keyframe-to-Pose (KF2P) generation, making precise spatiotemporal editing of signing sequences possible. Furthermore, \papername{} scales across four distinct sign languages, constituting the largest multilingual SLP framework to date, and integrates 3D Gaussian Splatting for photorealistic rendering. Extensive evaluations demonstrate that \papername{} achieves state-of-the-art across diverse SLP tasks and multilingual benchmarks. Our code is available at {\url{https://github.com/JianHe0628/SignSparK}}.
\end{abstract}

\section{Introduction}
\label{sec:intro}

Sign languages are linguistically rich, natural languages primarily used in Deaf communities, and are governed by precise hand shapes and fluid body dynamics~\cite{stokoe2001language,DianeSignCompare,Signvsspoken}. While Sign Language Translation (SLT) from video-to-text has been extensively studied~\cite{camgoz2018neural,zhou2023gloss,fish2025geo,sincan2025gloss,li2025uni,low2025sage}, its inverse, Sign Language Production (SLP)~\cite{saunders2020progressive,stoll2020text2sign,zelinka2020neural,baltatzis2024neural,zuo2024signs} has emerged only more recently, and carries its own unique set of structural challenges. Chief among these is the need to satisfy two demands at once: (i) each sign must be articulated with linguistic precision, and (ii) successive signs must connect with the smooth continuity of natural movement. Reconciling both characteristics within a single text-to-motion mapping has proven non-trivial, and existing methods thus sacrifice one for the other.

\begin{figure}[t]
    \centering
    \includegraphics[width=1\linewidth]{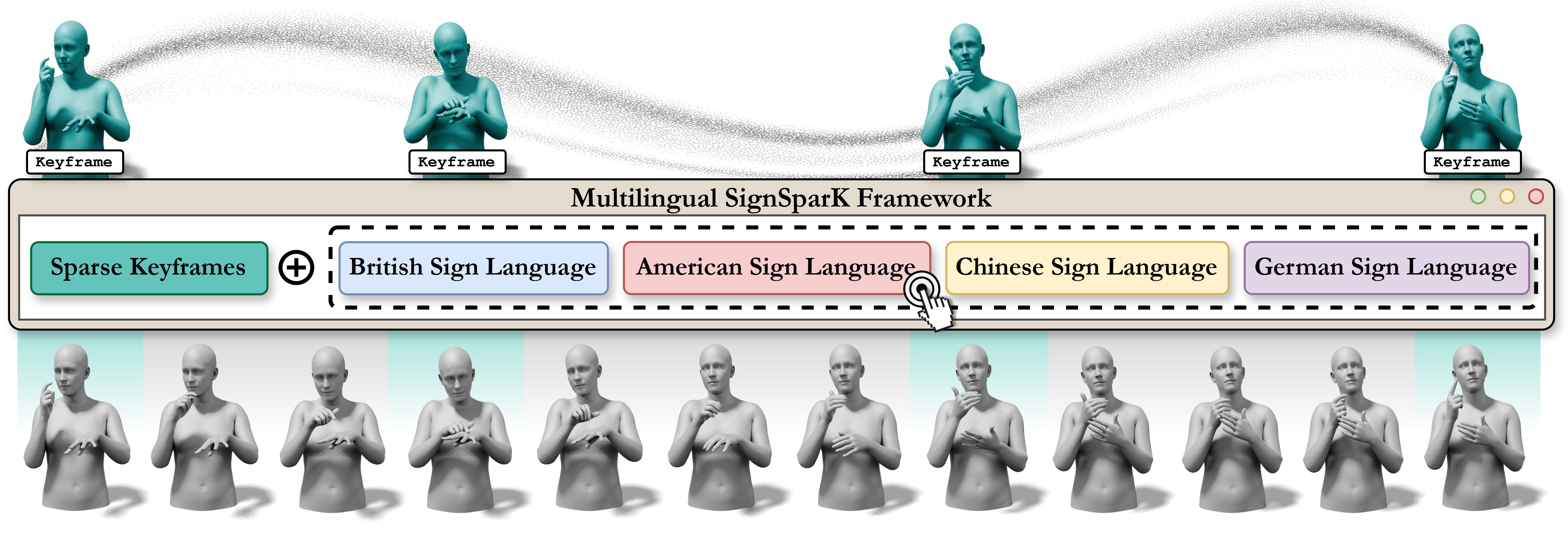}
    \caption{\textbf{\papername{}} is a Conditional Flow Matching model trained on sparse keyframes, and generates realistic and natural 3D signing avatars given spoken text. Designed for efficiency, \papername{} scales to four distinct sign languages under a unified framework.}
    \label{fig:teaser}
\end{figure}

For instance, \emph{direct} Text-to-Pose (T2P) models~\cite{tang2022gloss, zelinka2020neural, stoll2020text2sign,saunders2020progressive} attempt to regress this mapping directly, but often collapse under the modality gap, yielding under-articulated and unintelligible signing. Meanwhile, real-world T2P deployment instead favours sign dictionary-retrieval pipelines based on \emph{glosses}\footnote{Glosses are the written representations used to denote individual signs.}~\cite{stoll2018sign,saunders2022signing,zuo2024simple}, as text is first translated to a gloss sequence, before being retrieved from a pre-recorded gloss dictionary. This pipeline preserves articulation, but sacrifices fluency: the clips must still be stitched together, and current methods do so via \emph{naive interpolation}, producing signing motion that looks robotic. Additionally, current SLP paradigms are also bottlenecked by inaccuracies in monocular 3D estimation (e.g., HaMeR~\cite{pavlakos2024reconstructing}), as the ambiguity of lifting 2D views into 3D meshes yields pseudo-ground truth that propagates directly into trained models.

To address these challenges, we propose \papername{} (\textbf{Sign} Language Production with \textbf{Spar}se \textbf{K}eyframes), a large-scale SLP framework driven by a novel \emph{training paradigm}: learning to synthesize continuous signing sequences given spoken text and sparse keyframes as training inputs (\cref{fig:teaser}). Since the model must explicitly hit these sparse keyframes, they act as anchors that inherently prevent regression-to-the-mean. When applied at scale, \papername{} cannot rely on naive interpolation and must thus learn the underlying distribution of fluid signing instead. This makes \papername{} a drop-in replacement for the stitching stage of retrieval-based pipelines, substituting the hand-crafted interpolation methods with a learnt motion prior. Additionally, \papername{}'s training paradigm also uniquely unlocks the highly controllable task of Keyframe-to-Pose (KF2P) synthesis. Here, users can shift anchors to dictate signing speed, insert intermediates for signs with complex internal motion, or replace inaccurate keyframes with high-fidelity mocap data, bounding output quality directly to the conditioning signal rather than to imperfect pseudo-ground truth. Finally, \papername{}'s Flow Matching architecture also achieves high-quality synthesis in fewer than ten sampling steps, allowing us to scale to the largest multilingual SLP framework to date, encompassing German, Chinese, American, and British sign languages. 

However, keyframe annotations do not exist in current SLP datasets, rendering this training paradigm infeasible. To overcome this, we additionally introduce FAST, an extremely efficient sign language segmentation model designed to automatically mine linguistically meaningful temporal anchors at scale. By precisely identifying sign boundaries, FAST provides the foundational keyframe extractions required to train \papername{}. Beyond enabling \papername{}, FAST is also explicitly engineered to process large-scale corpora like BOBSL\cite{albanie2021bbc} with minimal computational overhead. It therefore offers immense standalone value for the broader community, as it can facilitate the rapid pseudo-annotation of emerging large-scale glossless datasets like YTSL-25 \cite{tanzer2024youtube} and CSL-News \cite{li2025uni}, and support downstream tasks such as sign spotting \cite{varol2022scaling} and tokenization \cite{low2025sage}.

By leveraging the keyframes extracted by FAST in training, \papername{} synthesizes articulate and naturalistic signing sequences directly within the 3D parametric spaces of SMPL-X~\cite{pavlakos2019expressive}, MANO~\cite{romero2017embodied}, and FLAME \cite{li2017learning}. Unlike prior 2D keypoint methods, this 3D formulation offers superior spatial depth and physical plausibility. We further show that 3D Gaussian Splatting (3DGS)~\cite{kerbl20233d} integrates naturally into our pipeline, overcoming the visual limitations of bare meshes by rendering the generated kinematics as photorealistic signing avatars. In short, the main contributions of our work can be summarized as follows:

\begin{itemize}
    \item \textbf{State-of-the-art segmentation.} We introduce FAST, an ultra-efficient sign segmentation model that achieves highly accurate boundary detection at scale and unlocks the linguistic keyframes required for our training paradigm.
    
    \item \textbf{Novel training paradigm.} We propose \papername{}, a generative SLP framework driven by sparse keyframe training. This approach overcomes under-articulation and ensures fluid motion while uniquely unlocking Keyframe-to-Pose (KF2P) synthesis for precise spatial and temporal control.
    
    \item \textbf{Unprecedented efficiency and scale.} We formulate a reconstruction-based Flow Matching objective for SLP and achieve a $100\times$ efficiency gain over prior models. We simultaneously leverage this efficiency to establish the largest multilingual SLP framework to date across four major sign languages.
    
    \item \textbf{State-of-the-art SLP performance.} Extensive evaluations demonstrate that \papername{} achieves state-of-the-art performance across a wide range of SLP regimes on multilingual sign language datasets.
    
\end{itemize}

\section{Related Work}
\label{sec:related_work}

\noindent\textbf{\emph{Sign Language Segmentation.}}
In continuous signing, the fluid transitional movements between consecutive signs, known as \emph{coarticulations}, create smooth motion continuity. Sign language segmentation seeks to temporally localize individual signs within these sequences, while disregarding coarticulatory transitions. Early approaches relied on statistical techniques \cite{5206523,5206647,5457527} and machine learning approaches \cite{Farag2019LearningMD}, but struggled with the multimodal, continuous nature of signing. Deep learning approaches later adopted sequential architectures such as the BiLSTM \cite{Bull2020AutomaticSO} and spatiotemporal models like the I3D \cite{8099985} to achieve notable improvements \cite{9413817,renz2021sign,9710309}. More recent studies, inspired by linguistics, then reframed sign segmentation as a Begin-In-Out (BIO) tagging task \cite{moryossef2023linguistically}. With Hands-On \cite{he2025hands}, advancing this task through strong 3D body \cite{ivashechkin2023improving} and hand \cite{pavlakos2024reconstructing} priors, achieving state-of-the-art on the DGS Corpus \cite{hanke-etal-2020-extending}. Our method, FAST, further improves this via a unimodal two-stream design and interpolated training, providing superior accuracy and significant efficiency gains.

\noindent\textbf{\emph{Sign Language Production.}}
Generating naturalistic signing videos or avatars from language or motion-based inputs defines the field of SLP. Early research was filled with graphics-based avatars driven by handcrafted linguistic rules and annotations~\cite{bangham2000virtual, cox2002tessa, efthimiou2012dicta, elghoul2011websign, zwitserlood2004synthetic}. Although linguistically structured, these pipelines produced unnatural motions, limiting acceptance within the Deaf community~\cite{kipp2011assessing}.

The advent of deep learning shifted SLP towards data-driven approaches. Owing to the non-monotonic mapping between sign and spoken languages, many systems introduced glosses as intermediate supervision, forming text-to-gloss-to-pose (T2G2P) frameworks~\cite{tang2022gloss, zelinka2020neural, stoll2020text2sign, saunders2020progressive, saunders2021continuous, saunders2021mixed, saunders2020adversarial}. In contrast, direct regression methods typically predict 2D/3D poses straight from text or linguistic inputs such as HamNoSys \cite{arkushin2023ham2pose} or glosses~\cite{chen2024semantic, huang2021towards, tang2025gloss}; however, they frequently produce under-articulated motion due to regression-to-the-mean effects. Alternatives like interpolating isolated signs~\cite{walshsign, saunders2022signing} improve said articulations, but rely on complex pipelines and smoothing filters that fail to capture natural human signing.

To enhance realism, methods employing GANs~\cite{stoll2018sign, stoll2020text2sign, saunders2022signing}, diffusion models~\cite{fang2025signdiff, wang2025advanced}, and recently 3DGS~\cite{ivashechkin2025signsplat} have also been explored. However, a growing trend now centres on 3D avatar generation, where models predict parametric human body models such as SMPL-X~\cite{pavlakos2019expressive} to enable full-body mesh reconstructions~\cite{baltatzis2024neural, yu2024signavatars, dong2024signavatar, zuo2024signs, bensabath2025text, symeonidis2026m3t}. These frameworks span diffusion~\cite{baltatzis2024neural, bensabath2025text}, VAE~\cite{dong2024signavatar}, and VQ-VAE~\cite{yu2024signavatars, zuo2024signs} paradigms, but they remain largely monolingual and computationally inefficient. To overcome this, Conditional Flow Matching (CFM) has recently emerged as a highly efficient generative alternative, with SignFlow~\cite{khan2025signflow} pioneering its use in SLP. However, its resulting synthesis still exhibits clear articulatory inaccuracies and its evaluation remains restricted to a single dataset.

Ultimately, progress in 3D production, multilingual scaling, and inference efficiency has remained highly fragmented. \papername{} instead presents a unified framework to bridge these domains. By integrating a keyframe-based paradigm for fluid articulation with a highly efficient reconstruction-based CFM formulation, as well as realistic rendering via 3DGS \cite{kerbl20233d}, we scale to a massive multilingual setting to provide a practical, high-fidelity system for the Deaf community.

\noindent\textbf{\emph{Human Body Motion Generation.}} 
Human motion generation is a long-standing problem in computer vision, conditioned on diverse modalities such as action labels~\cite{cai2018deep, yu2020structure, wang2020learning, petrovich2021action}, audio~\cite{shlizerman2018audio, lee2019dancing}, and increasingly, text~\cite{guo2022generating, petrovich2022temos, guo2022tm2t, tevet2022motionclip, guo2024momask, tevethuman, chen2023executing, zhang2024motiondiffuse}, enabled by large-scale datasets like KIT-ML~\cite{plappert2016kit}, BABEL~\cite{punnakkal2021babel}, and HumanML3D~\cite{guo2022generating}. Recent approaches typically represent motion as latent tokens within autoencoder frameworks for autoregressive Transformer-based generation \cite{petrovich2022temos, guo2024momask, tevet2022motionclip, zhang2023generating}, or leverage diffusion-based models~\cite{ho2020denoising} to produce temporally coherent sequences~\cite{tevethuman, shafir2024human, zhang2024motiondiffuse, karunratanakul2023guided, chen2023executing}.
However, despite significant progress, human motion generation approaches emphasize full-body motion, and thus neglect the fine-grained hand and finger dynamics essential for sign language production.

\noindent\textbf{\emph{Keyframe In-betweening Generation.}}
In motion/video synthesis, \emph{keyframe in-betweening} aims to generate intermediate frames given sparse keyframe constraints. Diffusion-based approaches have recently advanced this across domains, including video~\cite{wanggenerative, guo2024sparsectrl, jain2024video, xing2024dynamicrafter}, human motion~\cite{karunratanakul2023guided, tevethuman, cohan2024flexible}, and hand interactions~\cite{lin2025handdiffuse}.
However, existing human motion frameworks remain limited: MDM~\cite{tevethuman} suffers from foot-sliding and unrealistic transitions under keyframe conditioning; GMD~\cite{karunratanakul2023guided} handles sparse interpolation but focuses on pelvis trajectories; CondMDI~\cite{cohan2024flexible} is closest to our setting, but targets general locomotion without explicit hand modelling and relies on a computationally expensive diffusion process. In contrast, our approach enables efficient few-step sampling while explicitly capturing the fine-grained articulations essential for natural signing motion.

\section{Methodology}
This work proposes a sparse keyframe-based SLP training paradigm consisting of two primary components. Firstly, a sign language segmentation model (Sec. \ref{Sec:SignSeg}) that localizes sign boundaries and extracts linguistically meaningful keyframes; and secondly, a CFM model (Sec. \ref{Sec:SLG Flow}) that synthesizes smooth and realistic sign language motion conditioned on these sparse keyframes and spoken text input.

\subsection{Sign Language Segmentation} \label{Sec:SignSeg}

Sign language segmentation can be formulated as a \emph{BIO-tagging} problem, where each frame is classified as either \textbf{\texttt{B}} (beginning of a sign), \textbf{\texttt{I}} (inside a sign), or \textbf{\texttt{O}} (outside a sign), corresponding to the class indices $\{2, 1, 0\}$. In this setting, given an input video $\mathbf{X} \in \mathbb{R}^{T \times H \times W \times C}$, with $T$ frames, $H \times W$ spatial resolution, and $C$ channel dimensions, the task is to assign each frame with a one-hot label $\mathbf{Y} \in \mathbb{R}^{T \times 3}$, treating segmentation as a per-frame classification problem.

\noindent \textbf{\emph{Feature Representation.}} To capture signing dynamics, previous work relied on HaMeR-extracted MANO hand parameters~\cite{pavlakos2024reconstructing, romero2017embodied} and 3D pose skeletons~\cite{ivashechkin2023improving}, achieving state-of-the-art segmentation~\cite{he2025hands} but at high computational costs impractical for large-scale multilingual sign datasets. We instead adopt WiLoR~\cite{potamias2025wilor}, a MANO regression model that recently attained superior performance, while featuring a 45× faster and 32× more compact hand detector. In addition, we also introduce our segmentation model as a unimodal approach to reduce complexity and feature extraction time. This design choice was to specifically improve model efficiency, but interestingly yields no drop in performance (see supplementary).

\begin{figure}[t]
    \centering
    \includegraphics[width=1\linewidth]{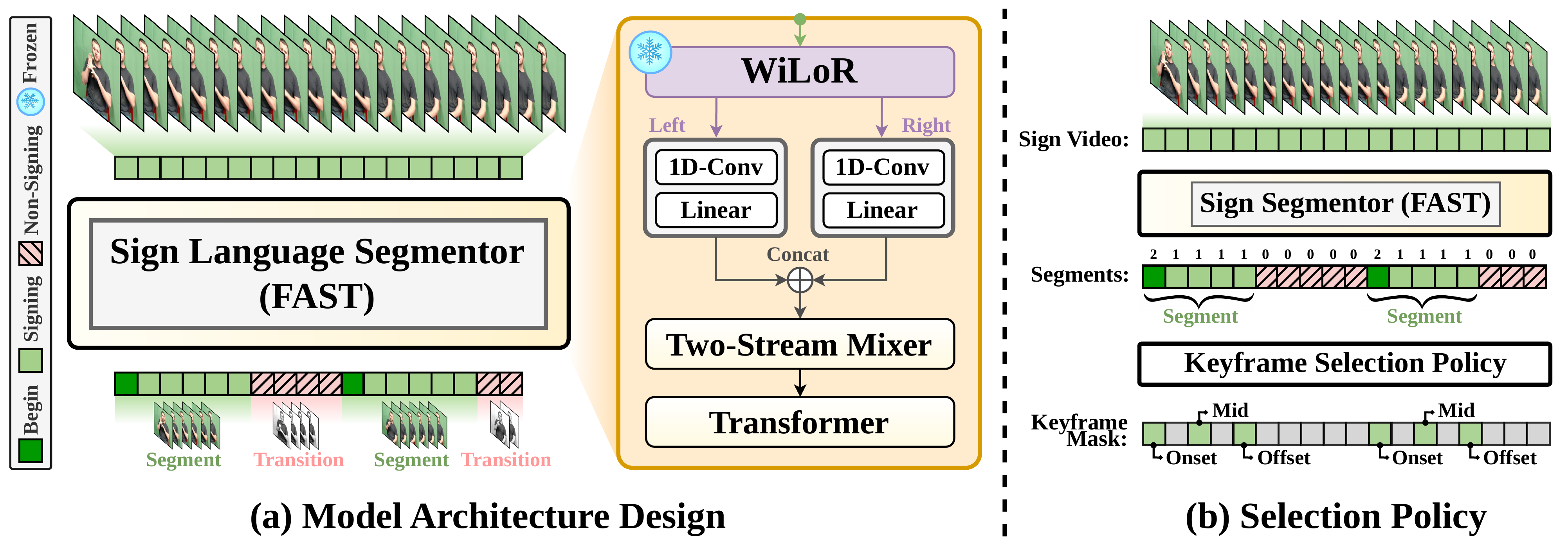}
    \caption{\textbf{Overview of FAST.} \textbf{(a) Architecture:} WiLoR first extracts the left and right hand representations from input frames. These are then encoded via parallel spatio-temporal streams, concatenated, and refined by a two-stream mixer before a Transformer generates dense per-frame BIO segmentation labels. \textbf{(b) Selection Policy:} Leveraging the predicted BIO segments, we explicitly isolate the onset, midpoint, and offset frames of each sign to construct a semantically rich keyframe mask.}
    \label{fig:sign_seg_overview}
\end{figure}

\noindent \textbf{\emph{Architecture Design.}}
Our \textbf{FAST} (\textbf{F}ast and \textbf{A}ccurate \textbf{S}ign segmen\textbf{T}ation) framework, is a transformer-based per-frame classification model (\cref{fig:sign_seg_overview}\textcolor{red}{a}) designed to capture hand shape transitions vital to segmentation. Specifically, given a video sequence $\bm{\mathbf{X}} = \{\mathbf{x}_1, \mathbf{x}_2, \dots, \mathbf{x}_T\}$,
we first extract per-frame MANO parameters for the left and right hands, $\mathbf{H}_L, \mathbf{H}_R \in \mathbb{R}^{T \times (J \times d_r)}$, where $J$ is the number of joints and $d_r$ the rotation dimensions. Here, we adopt 6D rotations, as it has demonstrated better suitability for deep learning~\cite{zhou2019continuity}. To capture the independent semantic contributions of each hand, FAST encodes left and right MANO features separately via parallel streams before fusing them to predict frame-wise BIO labels. We train the network using cross-entropy (frame-level) and CTC (sign-level) alongside temporal augmentations to yield a final segmentation model that is highly scalable, accurate, and computationally efficient. Further implementation details are provided in the supplementary material.

\noindent \textbf{\emph{Keyframe Selection Policy.}} Our selection policy aims to extract a sparse and semantically rich subset of frames that distils the motion and content of each sign. Leveraging the frame-wise predictions $y_t \in \{0,1,2\}$ from FAST, we ground our sampling in linguistic structure by explicitly isolating the exact temporal boundaries of each gesture: the onset ($y_t = 2$) and the offset (defined as $y_t = 1$ given $y_{t+1} = 0$). To capture the core content of the gesture, we further incorporate the segment midpoint. This resulting policy of onset$\rightarrow$mid$\rightarrow$offset (\cref{fig:sign_seg_overview}\textcolor{red}{b}) outperforms both random and heuristically augmented baselines (\cref{sec:ablation}), and also provides the optimal sparsity-performance trade-off (Supplementary).

\subsection{Sparse Keyframe-conditioned Training via Flow Matching} \label{Sec:SLG Flow}
\begin{figure}[t]
    \centering
    \includegraphics[width=1\linewidth]{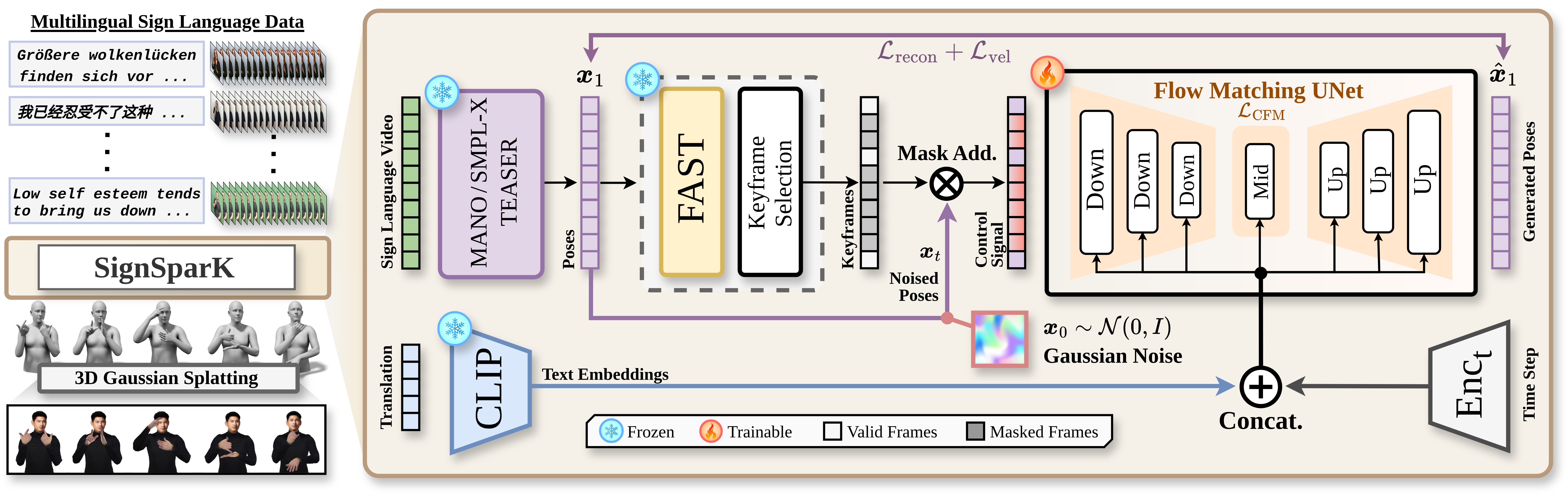}
    \caption{\textbf{Architecture of \papername{}}. (i) A sign language video is first processed by WiLoR, NLF, and TEASER to extract 3D parametric representations, while its text translation is embedded via Multilingual-CLIP \cite{carlsson2022cross}. (ii) FAST subsequently localizes sign segments, and the selection policy pinpoints the keyframes needed to form the control signal. (iii) A UNet, conditioned on timestep, control signal, and text, then reconstructs clean poses. (iv) These poses can then be rendered from meshes into realistic signing avatars via 3DGS. Further 3DGS implementation details in the supplementary}
    \label{fig:flowoverview}
\end{figure}

Unlike prior SLP training approaches that condition sign generation on spoken text~\cite{baltatzis2024neural}, tokenized motion sequences~\cite{zuo2024signs}, or textual motion descriptions~\cite{bensabath2025text}, we adopt the conditioning signal of sparse keyframes and spoken text as it enables the model to learn the underlying articulatory patterns of human signing, while also respecting its linguistic reference (Fig.~\ref{fig:flowoverview}). Formally, given a frame control signal $\mathbf{C} \in \mathbb{R}^{T \times D}$ and text condition $\mathcal{T}$, the objective is to synthesize a signing sequence $\mathbf{S} \in \mathbb{R}^{T \times D}$. The control signal $\mathbf{C}$ comprises of only $k$ keyframes ($k \ll T$), with the remaining $(T-k)$ frames corrupted with Gaussian noise $\mathcal{N}(\mathbf{0},\mathbf{I})$.

To represent signing sequence $\mathbf{S} \in \mathbb{R}^{T \times D}$, we regress SMPL-X upper-body parameters
$\mathbf{B} \in \mathbb{R}^{T \times (10 \times d_r)}$ using NLF~\cite{sarandi2024neural}, MANO hand parameters
$\mathbf{H}_L, \mathbf{H}_R \in \mathbb{R}^{T \times (15 \times d_r)}$ via WiLoR~\cite{potamias2025wilor}, and FLAME facial parameters $\mathbf{F} \in \mathbb{R}^{T \times 56}$ (50 expression coefficients and a 6D jaw rotation) with TEASER~\cite{teaser}.
We follow~\cite{zuo2024signs} by modeling the signing avatar as $10$ upper-body joints and $15$ joints per hand, but adopt 6D rotation parameterization~\cite{zhou2019continuity}.
Consistent with prior work, we also maintain $\mathbf{B}$, $(\mathbf{H}_L, \mathbf{H}_R)$, and additionally $\mathbf{F}$ as separate channels to preserve their distinct motion dynamics. Keyframes are then obtained using the FAST model and selection policy (Sec.~\ref{Sec:SignSeg}), resulting in the creation of binary mask $\mathcal{M} = \{m_1, m_2, \dots, m_T\}$, where $m_t = 1$ denotes a valid keyframe.

\noindent \textbf{\emph{Control Signal Construction.}}
Given the keyframe mask $\mathcal{M} = \{m_t\}_{t=1}^{T}$, we then construct the control signal to condition \papername{}. First, let $\vx_1 \in \mathbb{R}^{T \times d}$ denote the ground-truth pose sequence for the channels under consideration, where $d = 60$ for body and $d = 90$ for hands. Then, following the CFM formulation (preliminaries in supplementary), we define a continuous probability path between the original sequence $\vx_1$ and a noise sample $\vx_0$ via linear interpolation:

\begin{equation}
\vx_t = t\,\vx_1 + (1 - t)\,\vx_0, \quad
\vx_0 \sim \mathcal{N}(\mathbf{0}, \mathbf{I}),\quad t \in [0,1].
\end{equation}
The keyframe mask $\mathcal{M}$ is subsequently projected across the feature dimension and used to preserve the ground-truth poses at keyframe locations while injecting the interpolated noised poses elsewhere, yielding the control signal:
\begin{equation}
\mathbf{C} = \mathcal{M} \odot \vx_1 \;+\; (1 - \mathcal{M}) \odot \vx_t,
\end{equation}
where $\odot$ denotes element-wise multiplication. This sparsity forces the model to infer intermediate motion, promoting the learning of natural signing dynamics.

\noindent \textbf{\emph{Flow Matching Model.}}
To model the rich variability of human signing, we employ a conditional Flow Matching network $f_{\theta}$ that predicts a time-dependent vector field $\vv_t \in \mathbb{R}^{T \times d}$, guided by control signal $\mathbf{C}_t \in \mathbb{R}^{T \times d}$ and timestep $t$. Additionally, to enable T2P capabilities, the model is also conditioned on spoken translations $\mathcal{T}$, embedded via a text encoder $\vz_{\mathcal{T}} = \mathrm{Enc}(\mathcal{T})$, yielding the final conditional vector field parameterization $\vv_t = f_{\theta}(t,\mathbf{C}_t, \vz_{\mathcal{T}})$. Following CFM~\cite{liu2023flow}, the target vector field is defined as $\vu_t(\mathbf{C}_t \mid \vx_1) = \vx_1 - \mathbf{C}_t$, 
capturing the residual direction needed to move the noised control signal $\mathbf{C}_t$ towards the ground-truth $\vx_1$. The model is therefore trained to approximate this field by minimizing:

\begin{equation} \label{eq:cfmloss}
\mathcal{L}_{\mathrm{CFM}}(\theta)
= \mathbb{E}_{t,\,\mathbf{C}_t,\,\vx_1}
\big\| f_{\theta}(t, \mathbf{C}_t, \vz_{\mathcal{T}})
- (\vx_1 - \mathbf{C}_t) \big\|^2.
\end{equation}

To enable flexible inference with or without guidance inputs $\mathbf{C}_t$ and $\vz_{\mathcal{T}}$, we adopt classifier-free guidance (CFG)~\cite{ho2021classifier}; where during training, guidance inputs are randomly dropped with probability $\rho = 0.1$, and replaced by a null token $\varnothing$, teaching the model to handle both keyframe conditioned and unconditioned scenarios. At inference, the guidance's effect can then be modulated by interpolating the two predicted vector fields via $\vv_t^{(\gamma)} = \vv_t^{\text{uncond}} + \gamma \, (\vv_t^{\text{cond}} - \vv_t^{\text{uncond}})$, where \(\vv_t^{\text{cond}} = f_\theta(t, \mathbf{C}_t, \vz_{\mathcal{T}})\), \(\vv_t^{\text{uncond}} = f_\theta(t,\varnothing, \varnothing)\), and \(\gamma \ge 0\) controls the guidance scale. This formulation grants explicit control over the model's guidance adherence, and also directly enables end-to-end T2P translation at inference.

\noindent \textbf{\emph{Reconstruction Regularization.}}
While the CFM objective encourages the model to capture motion via vector fields, we observed that explicitly guiding the prediction towards the target pose improves convergence and sampling fidelity. To this end, we introduce a \emph{flow-based reconstruction loss}, which forces the vector field $\vv_t = f_{\theta}(t, \mathbf{C}_t, \vz_{\mathcal{T}})$ to recover the original pose $\vx_1$ from the noised control signal $\mathbf{C}_t$. Specifically, we estimate the original pose via a single-step Euler integration along the vector field from the current timestep $t$ towards $t=1$:
\begin{equation}
\vx_1^{\mathrm{est}} = \vx_t + (1-t)\, \vv_t, \quad t \in [0,1].
\end{equation}
We then supervise the network using mean-squared error to penalize significant deviations from the ground-truth poses:
\begin{equation}
\mathcal{L}_{\text{recon}} = 
\mathbb{E}_{\vx_1, \vx_t, t} \big\| \vx_1^{\text{est}} - \vx_1\big\|^2.
\end{equation}
Crucially, this reconstruction loss optimizes the vector field for \emph{one-step sampling}, enabling the extreme inference efficiency of \papername{}. To enforce temporal coherence and smooth dynamics, we further introduce a \emph{velocity-matching loss} that aligns the predicted inter-frame displacements with the ground truth:
\begin{equation}
\mathcal{L}_{\text{vel}} =
\mathbb{E}_{\vx_1, \vx_t, t}
\big\|
(\vx_{1,f+1}^{\text{est}} - \vx_{1,f}^{\text{est}})
- (\vx_{1,f+1} - \vx_{1,f})
\big\|^2,
\end{equation}
with \( f \) indexing the consecutive frames in the motion sequence. The final objective is then the combination of the CFM and reconstruction losses:
\begin{equation}
\mathcal{L}_{\text{SignSparK}} = 
\lambda_{\text{CFM}}\mathcal{L}_{\text{CFM}} 
+ \lambda_{\text{recon}}\mathcal{L}_{\text{recon}} 
+ \lambda_{\text{vel}}\mathcal{L}_{\text{vel}},
\end{equation}
where \(\lambda_{\text{recon}}\) and \(\lambda_{\text{vel}}\) weight the auxiliary terms. By explicitly reconstructing the target pose at each timestep via one-step integrations, \papername{} consequently learns more faithful and fluid signing motion, improving generation quality (\cref{sec:ablation}) while requiring far fewer sampling steps.

\section{Experiments} \label{Sec:experiments}
In this section, we evaluate both of our main contributions: the sign language segmentor \textbf{FAST} and the SLP framework \textbf{\papername{}}. We first outline the datasets and evaluation metrics used, followed by state-of-the-art comparisons, ablations, and qualitative results. As human evaluation remains the gold standard for assessing sign correctness, we also include a comprehensive user study.

\noindent \textbf{\emph{Datasets.}}
For sign segmentation, we follow standard practice~\cite{moryossef2020real,moryossef2023linguistically,he2025hands} and adopt MeineDGS~\cite{hanke-etal-2020-extending} as both our training and evaluation benchmark. This corpus offers precise frame-level boundary annotations for continuous signing, making it the only resource suitable for this task. Nonetheless, we show that FAST generalizes well even to unseen signers and datasets in the supplementary.

For SLP, we construct a multilingual training corpus following SOKE~\cite{zuo2024signs} by merging Phoenix14T~\cite{camgoz2018neural}, CSLDaily~\cite{zhou2021improving}, and How2Sign~\cite{duarte2021how2sign}. However, we further extend their setup by adding British Sign Language data from the large-scale BOBSL corpus~\cite{albanie2021bbc}, enhancing linguistic diversity and coverage. To prevent data dominance, only 10\% of BOBSL is used for training. For downstream evaluations, \papername{} is rigorously benchmarked across multiple SLP regimes. To ensure fair comparison with prior work, we align our test sets accordingly: (i) Sign Stitching (Gloss-to-Pose, G2P) is evaluated on Phoenix14T, MeineDGS, and BSLCorpus~\cite{Schembri2017BritishSL} following \cite{walshsign}, while (ii) the Text-to-Pose (T2P) and Keyframe-to-Pose (KF2P) tasks are assessed on Phoenix14T, CSLDaily, and How2Sign, following \cite{zuo2024signs}. Additional dataset statistics are provided in the supplementary.

\noindent \textbf{\emph{Evaluation Metrics.}} To evaluate segmentation performance, we follow the metrics adopted in prior work~\cite{moryossef2020real,moryossef2023linguistically,he2025hands}. Specifically, we report frame-level F1 score (F1) for BIO label accuracy, Intersection over Union (IoU) to measure temporal overlap between predicted and ground-truth segments, and the Segment Ratio (SR) to quantify over- or under-segmentation based on segment counts.

Meanwhile, to evaluate performance across SLP regimes, we employ task-specific protocols. For Sign Stitching, we follow~\cite{walshsign}, utilizing Dynamic Time Warping on joint positions (DTW-JPE) to evaluate motion coherence and a Back-Translation (B-T) model to measure semantic intelligibility. For the T2P and KF2P tasks, we adopt the protocol of SOKE~\cite{zuo2024signs}. Here, motion and spatial fidelity are measured via DTW errors on both Procrustes-aligned (DTW-PA-JPE) and unaligned (DTW-JPE) joint positions. For KF2P, we also report B-T BLEU-4 scores~\cite{papineni2002bleu} using the SL-Transformer model~\cite{camgoz2020sign}, consistent with most prior approaches~\cite{saunders2020progressive,huang2021towards,walshsign,saunders2021mixed,saunders2022signing}. However, since BLEU scores vary heavily based on B-T configurations, we additionally report \emph{relative} BLEU-4 scores, computed as the percentage drop from the ground-truth, for a more interpretable comparison. As for FLAME facial expressions, we evaluate them separately in \cref{facial_exp_eval}, as no prior SLP methods to our knowledge report comparable metrics.

\subsection{Sign Language Segmentation} \label{sec:FAST_results}

\begin{table*}[t]
    \centering
    \begin{minipage}[t]{0.364\linewidth}
        \centering
        \caption{We compare FAST against state-of-the-art sign segmentation models on the MeineDGS~\cite{hanke-etal-2020-extending} dataset.}
        \label{tab:sota_segmentor}
        \resizebox{\linewidth}{!}{%
        \begin{tabular}{@{}llccc@{}}
            \toprule
            \textbf{Method} & \textbf{Input Features} & \textbf{F1}$_\uparrow$ & \textbf{IoU}$_\uparrow$ & \textbf{SR} \\
            \midrule
            \rowcolor{gray!10}\multicolumn{5}{c}{\textsc{Single-Modality}} \\
            IO Tagger~\cite{moryossef2020real}            & Optical Flow    & --             & 0.460          & 1.090          \\
            BIO Tagger~\cite{moryossef2023linguistically} & 3D Body Pose    & 0.630          & 0.690          & 1.110          \\
            Hands-On~\cite{he2025hands}                   & HaMeR           & 0.830          & --             & --             \\
            \addlinespace
            \rowcolor{gray!10}\multicolumn{5}{c}{\textsc{Multi-Modality}} \\
            BIO Tagger~\cite{moryossef2023linguistically} & 3D Pose + Hand  & 0.590          & 0.630          & 1.130          \\
            Hands-On~\cite{he2025hands}                   & 3D Pose + HaMeR & 0.857          & 0.760          & 0.980          \\
            \midrule
            \rowcolor[HTML]{F0F8FF} \textbf{Ours (FAST)}  & \textbf{WiLoR}  & \textbf{0.860} & \textbf{0.772} & \textbf{1.010} \\
            \bottomrule
        \end{tabular}%
        }
    \end{minipage}%
    \hfill%
    \begin{minipage}[t]{0.62\linewidth}
        \centering
        \caption{We compare \papername{} to prior works on G2P and T2P stitching. We report absolute scores along with relative drops (B-T Score $\mid$ Drop \%) instead, as our B-T models yield higher BLEU-4 scores.}
        \label{tab:sota_stitch}
        \resizebox{\linewidth}{!}{%
        \begin{tabular}{@{}lcccccc@{}}
            \toprule
            \multirow{2}{*}{\textbf{Method}} &
            \multicolumn{2}{c}{\textbf{Phoenix-2014T}} &
            \multicolumn{2}{c}{\textbf{MeineDGS}} &
            \multicolumn{2}{c}{\textbf{BSLCorpus}} \\
            \cmidrule(lr){2-3} \cmidrule(lr){4-5} \cmidrule(l){6-7}
            & \textbf{DTW-J}$_\downarrow$ & \textbf{B-T}$_\uparrow$ $\mid$ Drop$_\downarrow$ &
              \textbf{DTW-J}$_\downarrow$ & \textbf{B-T}$_\uparrow$ $\mid$ Drop$_\downarrow$ &
              \textbf{DTW-J}$_\downarrow$ & \textbf{B-T}$_\uparrow$ $\mid$ Drop$_\downarrow$ \\
            \midrule
            \rowcolor{gray!10}
            \textit{SS GT}~\cite{walshsign} & \textit{0.00} & \textit{11.32} & \textit{0.00} & \textit{0.80} & \textit{0.00} & \textit{0.54} \\
            SS T2P (Iso)~\cite{walshsign}   & 59.40 & 3.12 $\mid$ \color{softpink}$\downarrow$72\% & 58.10 & 0.22 $\mid$ \color{softpink}$\downarrow$73\% & 58.80 & 0.28 $\mid$ \color{softpink}$\downarrow$41\% \\
            SS T2P (Cont)~\cite{walshsign}  & 57.20 & 6.67 $\mid$ \color{softpink}$\downarrow$48\% & 63.70 & 0.39 $\mid$ \color{softpink}$\downarrow$51\% & 57.50 & 0.41 $\mid$ \color{softpink}$\downarrow$24\% \\
            SS G2P (Iso)~\cite{walshsign}   & 59.30 & 2.49 $\mid$ \color{softpink}$\downarrow$78\% & -- & -- & -- & -- \\
            SS G2P (Cont)~\cite{walshsign}  & 58.70 & 5.95 $\mid$ \color{softpink}$\downarrow$47\% & -- & -- & -- & -- \\
            \midrule
            \rowcolor{gray!10}
            \textit{GT} & \textit{0.00} & \textit{14.77} & \textit{0.00} & \textit{0.99} & \textit{0.00} & \textit{1.23} \\
            \rowcolor[HTML]{F0F8FF} \textbf{\papername{}} &
            \textbf{9.01} & \textbf{11.70} $\mid$ \color{softpink}$\downarrow$\textbf{21\%} &
            \textbf{8.57} & \textbf{0.75} $\mid$ \color{softpink}$\downarrow$\textbf{24\%} &
            \textbf{5.24} & 1.13 $\mid$ \color{softpink}$\downarrow$8\% \\
            \bottomrule
        \end{tabular}%
        }
    \end{minipage}
\end{table*}

\noindent \textbf{\emph{State-of-the-Art Comparisons.}} In \cref{tab:sota_segmentor}, we compare our approach (FAST), against recent state-of-the-art (SOTA) segmentation models and demonstrate consistent improvements across all metrics, including F1 (+0.3\%), IoU (+1.2\%), and Segment Ratio, which is closer to the ideal value of 1 (1.01). Importantly, these gains are achieved using a unimodal setup, without relying on additional modalities. While the absolute improvements are modest, FAST is primarily designed for efficiency and scalability. In particular, FAST significantly outperforms Hands-On \cite{he2025hands} in speed, achieving a 45$\times$ acceleration in hand detection by leveraging WiLoR and an additional 2$\times$ speedup by avoiding 3D body pose extraction. Moreover, FAST also operates with substantially lower dimensionality, requiring $\mathbb{R}^{192}$ for WiLoR’s 6D rotations compared to Hands-On’s combined hand and body features of $\mathbb{R}^{288+104}$. Together, these characteristics afford us the accuracy, efficiency, and speed needed to scale segmentation to the large datasets used in this work. We show additional timing details in the supplementary.

\subsection{Sign Language Production}
\noindent \textbf{\emph{Sign Stitching Evaluation.}} Sign stitching is an SLP task that generates continuous motion by concatenating isolated glosses. In \cref{tab:sota_stitch}, we find that \papername{} outperforms all variants of the SOTA Sign Stitcher (SS)~\cite{walshsign}, achieving lower DTW-JPE across all datasets. Furthermore, minimal degradation from upper-bound B-T scores also confirms that our generated sequences preserve high linguistic fidelity. \papername{} notably achieves this superior performance while conditioning on only three keyframes per gloss, whereas SS relies on full-sequence pose inputs. Finally, our framework also exhibits robust zero-shot generalization, surpassing SS on both the unseen datasets of MeineDGS and BSLCorpus.

\begin{table}[t]
\centering
\caption{\textbf{Comparisons with state-of-the-art Text-to-Pose (T2P) models.} \papername{} consistently reduces both body and hand Joint Position Error (JPE) across all datasets in both gloss-free (GF) and sign-retrieval (SR) regimes. For SOKE~\cite{zuo2024signs}, we report its non-dictionary and final performances for the GF and SR settings, respectively. Results marked with $\dagger$ are reproduced by~\cite{zuo2024signs}. B-T metrics are excluded as the required evaluation model and ground-truth baselines are publicly unavailable, while How2Sign is omitted for SR-T2P evaluation due to its lack of gloss annotations.}
\label{tab:t2p_models}
\small
\setlength\tabcolsep{4pt}
\renewcommand{\arraystretch}{1.2}
\resizebox{\linewidth}{!}{%
\begin{tabular}{l | cccc | cccc | cccc}
\toprule
\multirow{3}{*}{\textbf{Method}} & \multicolumn{4}{c|}{\textbf{How2Sign}} & \multicolumn{4}{c|}{\textbf{CSLDaily}} & \multicolumn{4}{c}{\textbf{Phoenix-2014T}} \\
& \multicolumn{2}{c}{\textbf{DTW-PA-JPE}$_\downarrow$} & \multicolumn{2}{c|}{\textbf{DTW-JPE}$_\downarrow$} & \multicolumn{2}{c}{\textbf{DTW-PA-JPE}$_\downarrow$} & \multicolumn{2}{c|}{\textbf{DTW-JPE}$_\downarrow$} & \multicolumn{2}{c}{\textbf{DTW-PA-JPE}$_\downarrow$} & \multicolumn{2}{c}{\textbf{DTW-JPE}$_\downarrow$} \\
\cmidrule(lr){2-3} \cmidrule(lr){4-5} \cmidrule(lr){6-7} \cmidrule(lr){8-9} \cmidrule(lr){10-11} \cmidrule(lr){12-13}
& Body & Hand & Body & Hand & Body & Hand & Body & Hand & Body & Hand & Body & Hand \\
\midrule
\rowcolor{gray!10}\multicolumn{13}{c}{\textsc{Gloss-Free Text-to-Pose (GF-T2P)}} \\
\midrule
Prog. Trans.$^\dagger$ \cite{saunders2020progressive} & 14.15 & 11.57 & 14.74 & 30.17 & 15.98 & 12.91 & 16.30 & 32.63 & 13.67 & 11.95 & 15.01 & 31.77 \\
Text2Mesh$^\dagger$~\cite{stoll2022there}           & 13.99 & 13.47 & 15.50 & 32.97 & 13.47 & 12.10 & 13.76 & 30.37 & 13.48 & 12.06 & 14.04 & 31.64 \\
T2S-GPT$^\dagger$~\cite{yin2024t2s}                 & 11.48 & 6.39  & 12.65 & 18.44 & 11.94 & 5.93  & 12.32 & 15.43 & 10.38 & 6.47  & 11.65 & 19.09 \\
S-MotionGPT$^\dagger$ \cite{jiang2023motiongpt}     & 11.23 & 4.39  & 12.41 & 13.74 & 10.81 & 3.78  & 11.58 & 11.31 & 9.45  & 3.41  & 10.42 & 9.08  \\
SOKE (no dict.) \cite{zuo2024signs}                  & 7.91  & 3.10  & --    & --    & 7.58  & 2.17  & --    & --    & 6.16  & 1.85  & --    & --    \\
\rowcolor[HTML]{F0F8FF} \textbf{\papername{} (no KF)} & \textbf{7.26} & \textbf{2.72} & \textbf{6.30} & \textbf{11.43} & \textbf{7.26} & \textbf{2.00} & \textbf{6.27} & \textbf{10.63} & \textbf{5.24} & \textbf{1.52} & \textbf{4.40} & \textbf{7.10} \\
\midrule
\rowcolor{gray!10}\multicolumn{13}{c}{\textsc{Sign Retrieval Text-to-Pose (SR-T2P)}} \\
\midrule
SOKE \cite{zuo2024signs}                            & 6.82  & 2.35  & 7.75  & 10.08 & 6.24  & 1.71  & 7.38  & 9.68  & 4.77  & 1.38  & 6.04  & 7.72  \\
\rowcolor[HTML]{F0F8FF} \textbf{\papername{}}       & --    & --    & --    & --    & \textbf{4.87} & \textbf{1.37} & \textbf{4.89} & \textbf{6.56} & \textbf{3.68} & \textbf{1.18} & \textbf{3.51} & \textbf{4.82} \\
\bottomrule
\end{tabular}%
}
\end{table}

\begin{table}[t]
\centering
\caption{\textbf{Keyframe-to-Pose (KF2P) evaluation.} We compare \papername{} against a standard Spherical Linear Interpolation (SLERP) baseline, demonstrating that our data-driven approach outperforms traditional rotational interpolation on all metrics.}
\label{tab:kf2p_models}
\small
\setlength\tabcolsep{4.5pt}
\renewcommand{\arraystretch}{1.2}
\resizebox{\linewidth}{!}{%
\begin{tabular}{l | ccccc | ccccc | ccccc}
\toprule
\multirow{3}{*}{\textbf{Method}} & 
\multicolumn{5}{c|}{\textbf{How2Sign}} & 
\multicolumn{5}{c|}{\textbf{CSLDaily}} & 
\multicolumn{5}{c}{\textbf{Phoenix-2014T}} \\
& \multicolumn{2}{c}{DTW-PA-J$_\downarrow$} & \multicolumn{2}{c}{DTW-J$_\downarrow$} & \textbf{B-T}$_\uparrow$ \textbar\ Drop$_\downarrow$ &
\multicolumn{2}{c}{DTW-PA-J$_\downarrow$} & \multicolumn{2}{c}{DTW-J$_\downarrow$} & \textbf{B-T}$_\uparrow$ \textbar\ Drop$_\downarrow$ &
\multicolumn{2}{c}{DTW-PA-J$_\downarrow$} & \multicolumn{2}{c}{DTW-J$_\downarrow$} & \textbf{B-T}$_\uparrow$ \textbar\ Drop$_\downarrow$\\
\cmidrule(lr){2-3} \cmidrule(lr){4-5} \cmidrule(lr){6-6} 
\cmidrule(lr){7-8} \cmidrule(lr){9-10} \cmidrule(lr){11-11} 
\cmidrule(lr){12-13} \cmidrule(lr){14-15} \cmidrule(lr){16-16}
& Bdy & Hnd & Bdy & Hnd & B-4 \textbar\ \% & Bdy & Hnd & Bdy & Hnd & B-4 \textbar\ \% & Bdy & Hnd & Bdy & Hnd & B-4 \textbar\ \% \\
\midrule
\rowcolor{gray!10} \textit{GT} & \textit{0.00} & \textit{0.00} & \textit{0.00} & \textit{0.00} & \textit{3.37} & \textit{0.00} & \textit{0.00} & \textit{0.00} & \textit{0.00} & \textit{6.83} & \textit{0.00} & \textit{0.00} & \textit{0.00} & \textit{0.00} & \textit{14.77}\\
SLERP Baseline & 4.18 & 0.98 & 4.36 & 3.76 & 2.78 \textbar\ {\color{softpink}$\downarrow$18\%} & 4.39 & 0.85 & 4.05 & 4.35 & 4.88 \textbar\ {\color{softpink}$\downarrow$29\%} & 4.10 & 0.89 & 3.71 & 4.12 & 9.18 \textbar\ {\color{softpink}$\downarrow$38\%} \\
\rowcolor[HTML]{F0F8FF} \textbf{\papername{}} & \textbf{2.11} & \textbf{0.86} & \textbf{1.68} & \textbf{2.80} & \textbf{2.99} \textbar\ {\color{softpink}$\downarrow$\textbf{11\%}} & \textbf{2.50} & \textbf{0.63} & \textbf{2.02} & \textbf{2.71} & \textbf{5.85} \textbar\ {\color{softpink}$\downarrow$\textbf{14\%}} & \textbf{2.39} & \textbf{0.71} & \textbf{1.92} & \textbf{2.25} & \textbf{11.70} \textbar\ {\color{softpink}$\downarrow$\textbf{21\%}} \\
\bottomrule
\end{tabular}%
}
\end{table}

\noindent \textbf{\emph{T2P Evaluation.}} We evaluate \papername{} against SOTA T2P models in both gloss-free (GF) and sign-retrieval (SR) regimes (\cref{tab:t2p_models}). As GF-T2P requires the direct regression of continuous motion from spoken text, we evaluate \papername{} using zero-keyframe inference; a capability enabled by CFG. Notably, even in the complete absence of guiding spatial anchors, \papername{} still outperforms prior GF-T2P approaches across all metrics on all three datasets. This highlights that despite being optimized primarily via sparse conditioning, \papername{} still internalizes the complex linguistic mapping between text and signing kinematics.

While this confirms strong GF-T2P performance, retrieval-based SLP (SR-T2P, or T2G2P) is the regime we primarily target, as it grounds the signing in linguistically valid exemplars, guaranteeing articulation and understandability. We thus compare against SOKE's retrieval-augmented generation by adopting the T2G2P pipeline \cite{saunders2022signing, zuo2024simple}: a text-to-gloss (T2G) model first predicts the gloss sequence, which then indexes an isolated sign dictionary at inference. However, while prior methods retrieve \emph{full video clips} per gloss, we instead retrieve only \emph{three keyframes} per gloss (drastically more storage efficient), pad the keyframes at a fixed 8-frame interval and then jointly condition \papername{} with spoken text. We provide additional implementation details in the supplementary.

Shown in \cref{tab:t2p_models}, \papername{} significantly outperforms SOKE. On CSLDaily, we reduce body and hand DTW-PA-JPE to 4.87 ($\downarrow$22\%) and 1.37 ($\downarrow$20\%), respectively, with comparable reductions of 23\% and 14\% on Phoenix14T. Additionally, our reconstruction-based CFM also achieve unprecedented efficiency, accelerating inference by $100\times$ (10-step \papername{}'s 0.01s/vid vs. SOKE's 1.55s \cite{zuo2024signs}). These sampling steps can be further reduced to flexibly trade fidelity for speed.

\noindent \textbf{\emph{KF2P Evaluation.}} Lastly, we benchmark \papername{} against a SLERP baseline on the newly proposed KF2P regime (\cref{tab:kf2p_models}), and find clear performance gains across all datasets. This indicates that \papername{} successfully learnt the complex dynamics of signing rather than merely interpolating between sparse frames.

\begin{figure*}[t]
    \centering
    \begin{minipage}[t]{0.365\linewidth}
        \captionof{table}{Ablation study on FAST's architectural choices.}
        \label{tab:segmentablate}
        \footnotesize
        \resizebox{\linewidth}{!}{%
        \begin{tabular}{lcc}
            \toprule
            \textbf{Model Variant} & \textbf{F1}$_\uparrow$ & \textbf{IoU}$_\uparrow$ \\
            \midrule
            \rowcolor{gray!10} Single-Stream (baseline) & 0.848 & 0.744 \\
            \rowcolor[HTML]{FAF3E6} Two-Stream & 0.850 & 0.747 \\
            \quad + Full Temporal Res. (\textit{No Downsampling}) & 0.860 & 0.765 \\
            \rowcolor[HTML]{F0F8FF}\quad + Temporal Conv. (\textit{Kernel Size }$=3$) & \textbf{0.861} & \textbf{0.768} \\
            \quad + Dropout & 0.857 & 0.761 \\
            \bottomrule
        \end{tabular}%
        }\\[0.35em]
        \includegraphics[width=\linewidth]{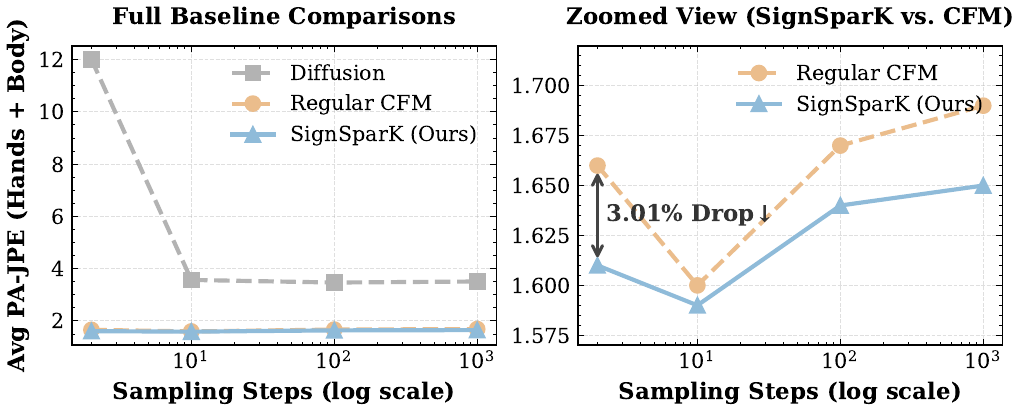}
        \captionof{figure}{Comparing \papername{} to diffusion and CFM models.}
        \label{fig:arch_compare}
    \end{minipage}
    \hfill
    \begin{minipage}[t]{0.622\linewidth}
        \captionof{table}{\textbf{Ablation on keyframe selection policy and loss configurations.} We evaluate the keyframe selection strategies and the contribution of each loss term towards body and hand reconstruction.}
        \label{tab:Spark_Ablate}
        \footnotesize
        \setlength{\tabcolsep}{4pt}
        \resizebox{\linewidth}{!}{%
        \begin{tabular}{lcccc}
        \toprule
        \multirow{2}{*}{\textbf{Ablation Experiment}} & \multicolumn{2}{c}{\textbf{Body}} & \multicolumn{2}{c}{\textbf{Hand}} \\
        \cmidrule(lr){2-3} \cmidrule(lr){4-5}
        & PA-MPJPE$_\downarrow$ & MPJPE$_\downarrow$ & PA-MPJPE$_\downarrow$ & MPJPE$_\downarrow$ \\
        \midrule
        \rowcolor{gray!10}\multicolumn{5}{c}{\textsc{Keyframe Selection Policy}} \\
        Random & $2.12_{\pm1.0}$ & $1.76_{\pm0.8}$ & $1.47_{\pm1.0}$ & $4.37_{\pm2.4}$ \\
        \rowcolor[HTML]{F0F8FF}\textbf{Segments Only} & $\mathbf{1.93}_{\pm0.8}$ & $\mathbf{1.61}_{\pm0.7}$ & $\mathbf{1.27}_{\pm0.5}$ & $\mathbf{3.94}_{\pm1.4}$ \\
        Segments w/ Random Masking & $2.02_{\pm0.9}$ & $1.68_{\pm0.7}$ & $1.31_{\pm0.5}$ & $4.11_{\pm1.4}$ \\
        \midrule
        \rowcolor{gray!10}\multicolumn{5}{c}{\textsc{Loss Configuration}} \\
        $\mathcal{L}_{\text{CFM}}$ & $1.93_{\pm0.8}$ & $1.61_{\pm0.7}$ & $1.27_{\pm0.5}$ & $3.97_{\pm1.4}$ \\
        $\mathcal{L}_{\text{recon}}$ & $2.02_{\pm0.9}$ & $1.68_{\pm0.7}$ & $1.27_{\pm0.5}$ & $4.09_{\pm1.4}$ \\
        $\mathcal{L}_{\text{vel}}$ & $8.43_{\pm2.2}$ & $7.12_{\pm1.9}$ & $1.51_{\pm0.5}$ & $9.60_{\pm2.7}$ \\
        $\mathcal{L}_{\text{CFM}} + \mathcal{L}_{\text{recon}}$ & $1.92_{\pm0.8}$ & $1.60_{\pm0.7}$ & $1.27_{\pm0.5}$ & $3.91_{\pm1.3}$ \\
        $\mathcal{L}_{\text{CFM}} + \mathcal{L}_{\text{vel}}$ & $1.95_{\pm0.8}$ & $1.63_{\pm0.7}$ & $1.27_{\pm0.5}$ & $4.01_{\pm1.4}$ \\
        \rowcolor[HTML]{F0F8FF}$\bm{\mathcal{L}_{\text{CFM}} + \mathcal{L}_{\text{recon}} + \mathcal{L}_{\text{vel}}}$ & $\mathbf{1.92}_{\pm0.8}$ & $\mathbf{1.60}_{\pm0.7}$ & $\mathbf{1.26}_{\pm0.5}$ & $\mathbf{3.91}_{\pm1.4}$ \\
        \bottomrule
        \end{tabular}%
        }
    \end{minipage}
\end{figure*}

\subsection{Ablation Studies} \label{sec:ablation}
\noindent \textbf{\emph{Sign Segmentor.}} In \cref{tab:segmentablate}, we ablate the design components of FAST. Using a single-stream design, consistent with prior work, yields an F1 of 0.85 and IoU of 0.74. Adopting a dual-stream architecture, where each stream processes one hand independently, then leads to minor gains in both metrics. Removing the 2× temporal downsampling used in \cite{he2025hands} further improves performance (F1 +1\%, IoU +1.8\%). Adding temporal convolution layers further boosts performances.

\noindent \textbf{\emph{Generative Architecture Comparison.}} \cref{fig:arch_compare} compares \papername{} against standard diffusion and CFM models across 1, 10, 100, and 1000 sampling steps. Here, the noise-based diffusion model severely underperforms in low-step regimes, requiring hundreds of steps to achieve competitive fidelity due to indirect noise optimization. Conversely, while \papername{} and standard CFM perform comparably at 10 steps, our framework demonstrates superior robustness elsewhere; notably yielding a 3\% gain in single-step generation. This stable fidelity retention stems directly from our reconstruction-based formulation, which explicitly penalizes pose prediction errors and enforces one-step sampling during training.

\noindent \textbf{\emph{Keyframe Selection Strategy.}} We then ablate the effectiveness of our segment-based keyframe selection policy by comparing it against random sampling and random segment masking during training (\cref{tab:Spark_Ablate}). Random sampling yields the weakest results, suggesting that unstructured frame selection disrupts the model’s ability to generate coherent signing motion. In contrast, our segment-based keyframes consistently yields strong performances across all metrics. Further adding random masking as augmentation then degrades results, indicating that the model learns better when trained directly on the structured anchors. Further ablations, such as joint masking, are provided in the supplementary.

\noindent \textbf{\emph{Loss Configuration.}} We analyze the effects of different loss combinations in \papername{} (\cref{tab:Spark_Ablate}). The baseline $\mathcal{L}_{\text{CFM}}$ already achieves strong performance, outperforming models trained with only $\mathcal{L}_{\text{recon}}$ or $\mathcal{L}_{\text{vel}}$. Adding $\mathcal{L}_{\text{recon}}$ further reduces joint errors, indicating complementary benefits between the CFM and reconstruction objectives. Combining all three losses then yields the best overall performance, with a 2:1:1 weighting of $(\mathcal{L}_{\text{CFM}}, \mathcal{L}_{\text{recon}}, \mathcal{L}_{\text{vel}})$ working best (see supplementary). We retain $\mathcal{L}_{\text{vel}}$ despite its small numerical contribution, as its role is qualitative: by penalizing inter-frame deviations from the pseudo-GT, it suppresses \emph{micro-jitters} that JPE-style metrics fail to capture. Overall, $\mathcal{L}_{\text{CFM}}$ lays a strong foundation, but the regularizations are what enforce the reconstruction accuracy and perceptual smoothness in \papername{}'s generated motion.

\begin{figure*}[t]
    \centering
    \begin{minipage}[t]{0.505\linewidth}
        \centering
        \captionof{table}{\textbf{Dataset and language token.} We ablate the contribution of multilingual training data and verify whether prepending language token identifiers (e.g., \texttt{<ASL>}) to text improves model performance.}
        \label{tab:datasetablation}
        \resizebox{\linewidth}{!}{%
        \begin{tabular}{@{}ccccccccc@{}}
            \toprule
            \textbf{Lang.} & \multicolumn{4}{c}{\textbf{Datasets}} & \multicolumn{2}{c}{\textbf{Body}} & \multicolumn{2}{c}{\textbf{Hand}} \\
            \cmidrule(lr){2-5} \cmidrule(lr){6-7} \cmidrule(l){8-9}
            \textbf{Tok.} & Ph-T & CSL & H2S & BSL & PA-JPE$_\downarrow$ & JPE$_\downarrow$ & PA-JPE$_\downarrow$ & JPE$_\downarrow$ \\
            \midrule
            \xmark & \cmark & \xmark & \xmark & \xmark & $2.87_{\pm1.2}$ & $2.49_{\pm1.0}$ & $1.51_{\pm0.5}$ & $5.20_{\pm1.9}$ \\
            \xmark & \cmark & \cmark & \xmark & \xmark & $2.66_{\pm1.1}$ & $2.29_{\pm0.9}$ & $1.36_{\pm0.5}$ & $4.58_{\pm1.6}$ \\
            \rowcolor[HTML]{FAF3E6}\xmark & \cmark & \cmark & \cmark & \xmark & $1.92_{\pm0.8}$ & $1.61_{\pm0.7}$ & $1.27_{\pm0.5}$ & $3.90_{\pm1.4}$ \\
            \midrule
            \xmark & \cmark & \cmark & \cmark & \cmark & $1.93_{\pm0.8}$ & $1.61_{\pm0.7}$ & $1.27_{\pm0.5}$ & $3.94_{\pm1.4}$ \\
            \rowcolor[HTML]{F0F8FF}\cmark & \cmark & \cmark & \cmark & \cmark & $\mathbf{1.90}_{\pm0.8}$ & $\mathbf{1.59}_{\pm0.7}$ & $\mathbf{1.27}_{\pm0.5}$ & $\mathbf{3.90}_{\pm1.4}$ \\
            \bottomrule
        \end{tabular}%
        }
    \end{minipage}%
    \hfill%
    \begin{minipage}[t]{0.475\linewidth}
        \centering
        ~\\[1\baselineskip]
        \includegraphics[width=\linewidth]{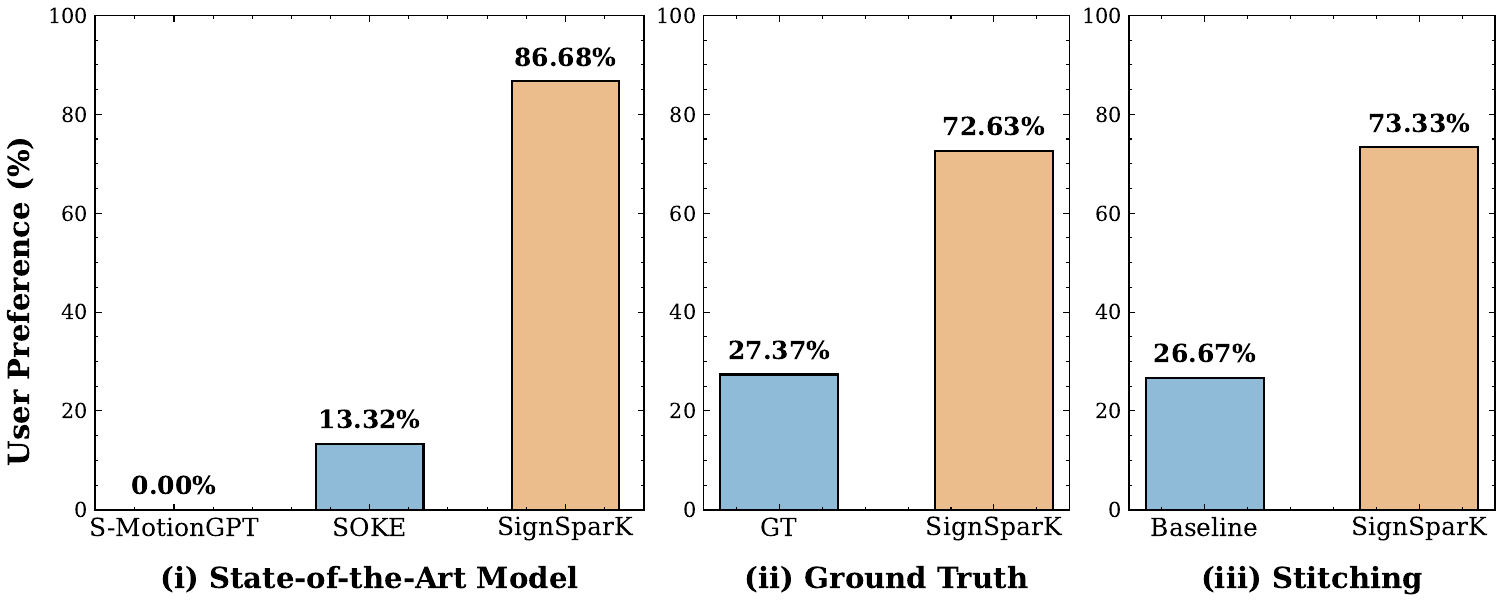}
        \captionof{figure}{\textbf{\papername{} user study.} We conduct a study with \textit{6 Deaf and 10 hearing BSL signers}, comparing \papername{} against SOTA and baseline models.}
        \label{fig:userstudy}
    \end{minipage}
\end{figure*}

\noindent \textbf{\emph{Contribution of Datasets.}} In \cref{tab:datasetablation}, we find that training \papername{} on more multilingual datasets generally improves downstream performance. However, this trend was initially disrupted by the inclusion of BOBSL, which increased overall error rates. Interestingly, introducing a language identifier token then resolves this issue, suggesting that the model was confusing American and British Sign Languages. Since both shared English as a spoken language, the model's ability to produce language-specific signing motion was likely hindered.

\noindent \textbf{\emph{User Study.}} Beyond quantitative metrics, we also conducted a forced-choice user study with 16 signers, evaluating perceived naturalness and visual alignment (\cref{fig:userstudy}). Across three distinct scenarios, \papername{} was overwhelmingly preferred. Against SOTA baselines (SOKE and S-MotionGPT), our model was favored in 86.68\% of trials. Interestingly, \papername{} was even preferred over pseudo-ground-truth extractions in 72.63\% of cases, as our learned signing prior significantly reduced the temporal jitter inherent to frame-wise 3D estimators. Finally, for sign stitching, our keyframe-conditioned generation also outperformed standard interpolation (73.33\% preference) due to its more fluid coarticulation. Comprehensive study details are provided in the supplementary material.

\begin{figure*}[t]
    \centering
        \includegraphics[width=1\linewidth]{fig/SignSpark_Qual_Results.jpg}
        \caption{\textbf{Qualitative Results.} \papername{} exhibits strong visual results across diverse tasks and sign languages. Top: KF2P predictions on BOBSL. Bottom-left: SOTA T2P Comparisons against SOKE~\cite{zuo2024signs} on CSL-Daily and Phoenix14T. Bottom-right: Photorealism comparison between \papername{} with 3DGS and SignGAN~\cite{saunders2022signing} on How2Sign.}
    \label{fig:qual_results}
\end{figure*}

\subsection{Qualitative Results}
In \cref{fig:qual_results}, we present qualitative examples on all four training datasets. On the challenging BOBSL dataset, \papername{} was able to faithfully reproduce body posture, hand positions, and common handshapes such as open palms and pointing gestures when given keyframe anchors. While occasional inaccuracies, such as incomplete hand closures, were observed, these were often caused by imprecisions in the ground-truth SMPL-X and MANO reconstructions provided by current single-view extraction approaches. On T2P SLP, \papername{} demonstrates clear advantages over SOKE~\cite{zuo2024signs}, as we find that \papername{} achieves consistently accurate signing with coherent articulations across consecutive signs. To render photorealistic avatars, we pair \papername{} with HuGeDiff~\cite{ivashechkin2025hugediff} due to its lightweight design, but note that \papername{} is inherently compatible with any SMPL-based 3DGS pipeline (e.g., GUAVA~\cite{zhang2025guava}). As seen in \cref{fig:qual_results}, the 3DGS renderings preserve precise hand geometry and yield more intelligible signing than GAN-based pipelines such as SignGAN~\cite{saunders2022signing}, which struggle to model fine-grained bimanual interactions. Further video comparisons are also provided in the supplementary to better demonstrate \papername{}'s generative performance.

\subsection{Facial Expression Evaluation} \label{facial_exp_eval}
\begin{figure*}[t]
\centering
\begin{minipage}[t]{0.75\linewidth}
    \centering
    \raisebox{-87pt}{\includegraphics[width=\linewidth]{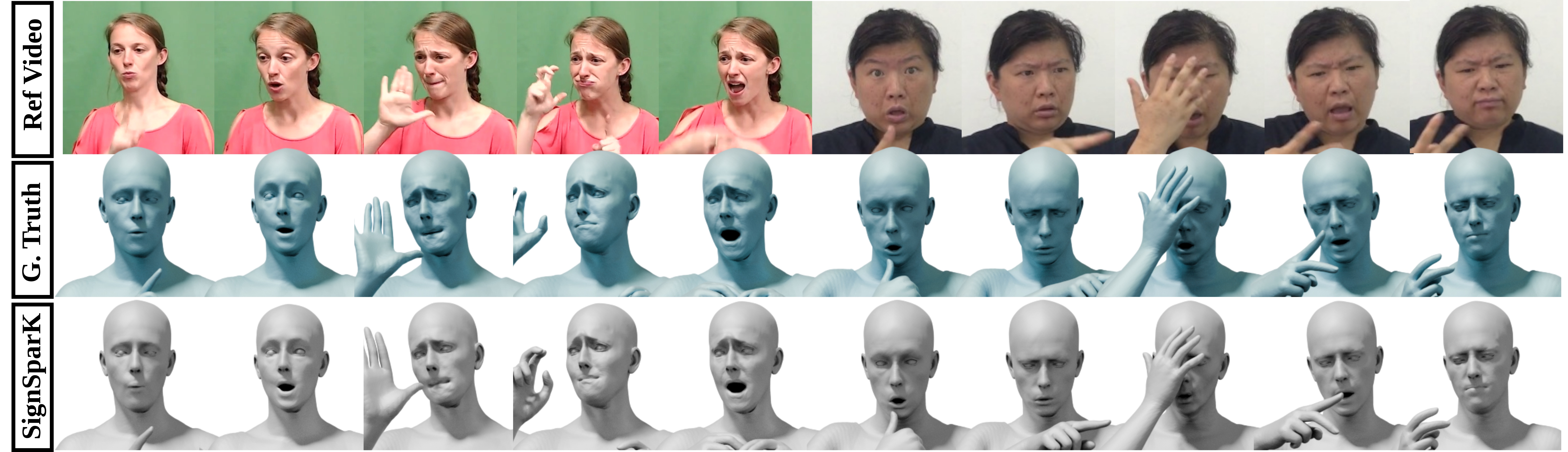}}
    \captionof{figure}{\textbf{Facial Expression Comparison.} We provide close-up comparisons between \papername{}'s synthesised facial expressions against the pseudo-ground truth from TEASER \cite{teaser} and confirm that our formulation transfers seamlessly to FLAME parameters.}
    \label{fig:facecompare}
\end{minipage}%
\hfill%
\begin{minipage}[t]{0.23\linewidth}
    \centering
    \captionof{table}{\textbf{Facial PA-VPE} values on How2Sign and CSL-Daily. Reported as a standalone reference as no comparable FLAME-based SLP baselines exist.}
    \label{tab:face_pavpe}
    \resizebox{\linewidth}{!}{%
    \begin{tabular}{@{}lc@{}}
        \toprule
        \textbf{Dataset} & \textbf{PA-VPE}$_\downarrow$ \\
        \midrule
        How2Sign  & $0.07_{\pm 0.02}$ \\
        CSL-Daily & $0.06_{\pm 0.02}$ \\
        \bottomrule
    \end{tabular}}
\end{minipage}
\end{figure*}

Non-manual features, and facial expressions in particular, carry essential grammatical and prosodic information in sign language~\cite{pfaunonmanuals}, yet to our knowledge, prior SMPL-based SLP methods have rarely, if ever, modelled them explicitly. Given how central facial articulations are to comprehension, we considered its inclusion in \papername{} essential. Identifying a feedforward extraction pipeline of sufficient fidelity for the subtle expressions characteristic of sign language proved a persistent challenge, but we ultimately adopted TEASER~\cite{teaser} for its balance of scalability and reconstruction quality, and used it to extract FLAME parameters across our datasets. As shown in \cref{fig:facecompare}, training \papername{} on these extractions yields tractable expressions that match both the reference video and the pseudo-GT, despite being synthesised from only sparse keyframes. As facial expression metrics have, to our knowledge, not been reported on standard SLP benchmarks, we provide \papername{}'s PA-VPE$_\text{face}$ as a standalone reference, measuring 0.07 on How2Sign and 0.06 on CSL-Daily (\cref{tab:face_pavpe}). We hope these initial numbers can serve as a starting point for future SMPL-based SLP methods.

\section{Conclusions}
This paper presents a unified framework for large-scale SLP built upon two core contributions. First, we introduce FAST, an ultra-efficient segmentor that establishes a new SOTA and enables the massive-scale extraction of linguistic keyframes. Second, we propose \papername{}, a CFM model driven by a novel training paradigm: synthesizing high-fidelity 3D signing motion directly from sparse keyframes. Coupled with a reconstruction-based objective, \papername{} achieves a >$100\times$ efficiency gain over prior methods. This efficiency directly enables our expansion across ASL, BSL, CSL, and DGS, establishing the largest multilingual SLP framework to date. Finally, by integrating 3DGS, \papername{} also overcomes the limitations of bare meshes to render photorealistic, identity-diverse avatars. Supported by extensive evaluations demonstrating SOTA kinematic fidelity and robust generalization, this work equips the community with a highly scalable segmentation tool and a multilingual generative prior for diverse SLP tasks.

\section*{Acknowledgements}
This work was supported by EPSRC grant APP24554 (SignGPT-EP/Z535370/1), EPSRC grant APP78083 (UMCS UKRI3927), as well as through funding from Google.org via the AI for Global Goals scheme. The authors acknowledge the use of Isambard-AI National AI Research Resource (AIRR) funded by UK DSIT via UKRI and STFC [ST/AIRR/I-A-I/1023]. Jianhe Low additionally acknowledges a bursary from the Rabin Ezra Scholarship Trust. This work reflects only the authors' views and the funders are not responsible for any use that may be made of the information it contains. 

%
%
\bibliographystyle{splncs04}
\bibliography{main}

\begin{thebibliography}{100}
\providecommand{\url}[1]{\texttt{#1}}
\providecommand{\urlprefix}{URL }
\providecommand{\doi}[1]{https://doi.org/#1}

\bibitem{albanie2021bbc}
Albanie, S., Varol, G., Momeni, L., Bull, H., Afouras, T., Chowdhury, H., Fox, N., Woll, B., Cooper, R., McParland, A., Zisserman, A.: {BOBSL: BBC-O}xford british sign language dataset. arXiv preprint arXiv:2111.03635  (2021)

\bibitem{arkushin2023ham2pose}
Arkushin, R.S., Moryossef, A., Fried, O.: Ham2pose: Animating sign language notation into pose sequences. In: Proceedings of the IEEE/CVF Conference on Computer Vision and Pattern Recognition (CVPR). pp. 21046--21056 (2023)

\bibitem{baltatzis2024neural}
Baltatzis, V., Potamias, R.A., Ververas, E., Sun, G., Deng, J., Zafeiriou, S.: Neural sign actors: A diffusion model for 3d sign language production from text. In: Proceedings of the IEEE/CVF Conference on Computer Vision and Pattern Recognition (CVPR). pp. 1985--1995 (2024)

\bibitem{bangham2000virtual}
Bangham, J.A., Cox, S., Elliott, R., Glauert, J.R., Marshall, I., Rankov, S., Wells, M.: Virtual signing: Capture, animation, storage and transmission-an overview of the visicast project. In: IEE Seminar on Speech and Language Processing for Disabled and Elderly People (Ref. No. 2000/025). pp.~6--1. IET (2000)

\bibitem{bensabath2025text}
Bensabath, L., Petrovich, M., Varol, G.: Text-driven 3d hand motion generation from sign language data. In: Proceedings of the IEEE/CVF Conference on Computer Vision and Pattern Recognition (CVPR). pp. 23095--23105 (2026)

\bibitem{5206523}
Buehler, P., Zisserman, A., Everingham, M.: Learning sign language by watching tv (using weakly aligned subtitles). In: Proceedings of the IEEE/CVF Conference on Computer Vision and Pattern Recognition (CVPR). pp. 2961--2968 (2009)

\bibitem{9710309}
Bull, H., Afouras, T., Varol, G., Albanie, S., Momeni, L., Zisserman, A.: Aligning subtitles in sign language videos. In: Proceedings of the IEEE/CVF International Conference on Computer Vision (ICCV). pp. 11532--11541 (2021)

\bibitem{Bull2020AutomaticSO}
Bull, H., Gouiff{\`e}s, M., Braffort, A.: Automatic segmentation of sign language into subtitle-units. In: European Conference on Computer Vision Workshops (ECCVW). pp. 186--198. Springer (2020)

\bibitem{cai2018deep}
Cai, H., Bai, C., Tai, Y.W., Tang, C.K.: Deep video generation, prediction and completion of human action sequences. In: Proceedings of the European Conference on Computer Vision (ECCV). pp. 366--382 (2018)

\bibitem{camgoz2018neural}
Camgoz, N.C., Hadfield, S., Koller, O., Ney, H., Bowden, R.: Neural sign language translation. In: Proceedings of the IEEE/CVF Conference on Computer Vision and Pattern Recognition (CVPR). pp. 7784--7793 (2018)

\bibitem{camgoz2020sign}
Camgoz, N.C., Koller, O., Hadfield, S., Bowden, R.: Sign language transformers: Joint end-to-end sign language recognition and translation. In: Proceedings of the IEEE/CVF Conference on Computer Vision and Pattern Recognition (CVPR). pp. 10023--10033 (2020)

\bibitem{carlsson2022cross}
Carlsson, F., Eisen, P., Rekathati, F., Sahlgren, M.: Cross-lingual and multilingual clip. In: International Conference on Language Resources and Evaluation (LREC). pp. 6848--6854 (2022)

\bibitem{8099985}
Carreira, J., Zisserman, A.: Quo vadis, action recognition? a new model and the kinetics dataset. In: Proceedings of the IEEE/CVF Conference on Computer Vision and Pattern Recognition (CVPR). pp. 4724--4733 (2017)

\bibitem{chen2024semantic}
Chen, S., Wang, Q., Wang, Q.: Semantic-driven diffusion for sign language production with gloss-pose latent spaces alignment. Computer Vision and Image Understanding (CVIU)  \textbf{246},  104050 (2024)

\bibitem{chen2023executing}
Chen, X., Jiang, B., Liu, W., Huang, Z., Fu, B., Chen, T., Yu, G.: Executing your commands via motion diffusion in latent space. In: Proceedings of the IEEE/CVF Conference on Computer Vision and Pattern Recognition (CVPR). pp. 18000--18010 (2023)

\bibitem{cohan2024flexible}
Cohan, S., Tevet, G., Reda, D., Peng, X.B., van~de Panne, M.: Flexible motion in-betweening with diffusion models. In: Special Interest Group on Computer Graphics and Interactive Techniques (SIGGRAPH). pp.~1--9. ACM (2024)

\bibitem{5206647}
Cooper, H., Bowden, R.: Learning signs from subtitles: A weakly supervised approach to sign language recognition. In: Proceedings of the IEEE/CVF Conference on Computer Vision and Pattern Recognition (CVPR). pp. 2568--2574 (2009)

\bibitem{Schembri2017BritishSL}
Cormier, K., Fenlon, J., Gulamani, S., Smith, S.: Bsl corpus annotation conventions (2017), \url{https://bslcorpusproject.org/wp-content/uploads/BSLCorpus_AnnotationConventions_v3.0_-March2017.pdf}, accessed: 2026-06-24

\bibitem{cox2002tessa}
Cox, S., Lincoln, M., Tryggvason, J., Nakisa, M., Wells, M., Tutt, M., Abbott, S.: Tessa, a system to aid communication with deaf people. In: Proceedings of the International ACM Conference on Assistive Technologies (ASSETS). pp. 205--212 (2002)

\bibitem{dong2024signavatar}
Dong, L., Chaudhary, L., Xu, F., Wang, X., Lary, M., Nwogu, I.: Signavatar: Sign language 3d motion reconstruction and generation. In: IEEE International Conference on Automatic Face and Gesture Recognition (FG). pp. 1--10. IEEE (2024)

\bibitem{duarte2021how2sign}
Duarte, A., Palaskar, S., Ventura, L., Ghadiyaram, D., DeHaan, K., Metze, F., Torres, J., Giro-i Nieto, X.: How2sign: A large-scale multimodal dataset for continuous american sign language. In: Proceedings of the IEEE/CVF Conference on Computer Vision and Pattern Recognition (CVPR). pp. 2735--2744 (2021)

\bibitem{efthimiou2012dicta}
Efthimiou, E., Fotinea, S.E., Hanke, T., Glauert, J., Bowden, R., Braffort, A., Collet, C., Maragos, P., Lefebvre-Albaret, F.: The dicta-sign wiki: Enabling web communication for the deaf. In: International Conference on Computers for Handicapped Persons (ICCHP). pp. 205--212. Springer (2012)

\bibitem{elghoul2011websign}
ElGhoul, O., Jemni, M.: Websign: A system to make and interpret signs using 3d avatars. In: Proceedings of the International Workshop on Sign Language Translation and Avatar Technology (SLTAT). vol.~23 (2011)

\bibitem{fang2025signdiff}
Fang, S., Sui, C., Zhou, Y., Zhang, X., Zhong, H., Tian, Y., Chen, C.: Signdiff: Diffusion model for american sign language production. In: IEEE International Conference on Automatic Face and Gesture Recognition (FG). pp. 1--11. IEEE (2025)

\bibitem{Farag2019LearningMD}
Farag, I., Brock, H.: Learning motion disfluencies for automatic sign language segmentation. In: IEEE International Conference on Acoustics, Speech and Signal Processing (ICASSP). pp. 7360--7364 (2019)

\bibitem{fish2025geo}
Fish, E., Bowden, R.: Geo-sign: Hyperbolic contrastive regularisation for geometrically aware sign language translation. In: Advances in Neural Information Processing systems (NeurIPS) (2025)

\bibitem{guo2024momask}
Guo, C., Mu, Y., Javed, M.G., Wang, S., Cheng, L.: Momask: Generative masked modeling of 3d human motions. In: Proceedings of the IEEE/CVF Conference on Computer Vision and Pattern Recognition (CVPR). pp. 1900--1910 (2024)

\bibitem{guo2022generating}
Guo, C., Zou, S., Zuo, X., Wang, S., Ji, W., Li, X., Cheng, L.: Generating diverse and natural 3d human motions from text. In: Proceedings of the IEEE/CVF Conference on Computer Vision and Pattern Recognition (CVPR). pp. 5152--5161 (2022)

\bibitem{guo2022tm2t}
Guo, C., Zuo, X., Wang, S., Cheng, L.: Tm2t: Stochastic and tokenized modeling for the reciprocal generation of 3d human motions and texts. In: Proceedings of the European Conference on Computer Vision (ECCV). pp. 580--597. Springer (2022)

\bibitem{guo2024sparsectrl}
Guo, Y., Yang, C., Rao, A., Agrawala, M., Lin, D., Dai, B.: Sparsectrl: Adding sparse controls to text-to-video diffusion models. In: Proceedings of the European Conference on Computer Vision (ECCV). pp. 330--348. Springer (2024)

\bibitem{ho2020denoising}
Ho, J., Jain, A., Abbeel, P.: Denoising diffusion probabilistic models. Advances in Neural Information Processing systems (NeurIPS)  \textbf{33},  6840--6851 (2020)

\bibitem{ho2021classifier}
Ho, J., Salimans, T.: Classifier-free diffusion guidance. In: Advances in Neural Information Processing systems (NeurIPS) Workshop (2021)

\bibitem{huang2021towards}
Huang, W., Pan, W., Zhao, Z., Tian, Q.: Towards fast and high-quality sign language production. In: Proceedings of the ACM International Conference on Multimedia (MM). pp. 3172--3181. ACM (2021)

\bibitem{ivashechkin2023improving}
Ivashechkin, M., Mendez, O., Bowden, R.: Improving 3d pose estimation for sign language. In: IEEE International Conference on Acoustics, Speech and Signal Processing Workshops (ICASSPW). pp.~1--5. IEEE (2023)

\bibitem{ivashechkin2025hugediff}
Ivashechkin, M., Mendez, O., Bowden, R.: {HuGeDiff: 3D Human Generation via Diffusion with Gaussian Splatting}. In: British Machine Vision Conference (BMVC). BMVA (2025)

\bibitem{ivashechkin2025signsplat}
Ivashechkin, M., Mendez, O., Bowden, R.: Signsplat: Rendering sign language via gaussian splatting. arXiv preprint arXiv:2505.02108  (2025)

\bibitem{jain2024video}
Jain, S., Watson, D., Tabellion, E., Poole, B., Kontkanen, J., et~al.: Video interpolation with diffusion models. In: Proceedings of the IEEE/CVF Conference on Computer Vision and Pattern Recognition (CVPR). pp. 7341--7351 (2024)

\bibitem{jiang2023motiongpt}
Jiang, B., Chen, X., Liu, W., Yu, J., Yu, G., Chen, T.: Motiongpt: Human motion as a foreign language. Advances in Neural Information Processing systems (NeurIPS)  \textbf{36},  20067--20079 (2023)

\bibitem{karunratanakul2023guided}
Karunratanakul, K., Preechakul, K., Suwajanakorn, S., Tang, S.: Guided motion diffusion for controllable human motion synthesis. In: Proceedings of the IEEE/CVF International Conference on Computer Vision (ICCV). pp. 2151--2162 (2023)

\bibitem{kerbl20233d}
Kerbl, B., Kopanas, G., Leimk{\"u}hler, T., Drettakis, G.: 3d gaussian splatting for real-time radiance field rendering. In: Transactions on Graphics (TOG). vol.~42, pp. 139--1. ACM (2023)

\bibitem{khan2025signflow}
Khan, N., Wu, B., Tan, S., Ishi, C.T., Nakadai, K.: Signflow: End-to-end sign language generation for one-to-many modeling using conditional flow matching. In: Proceedings of the 27th International Conference on Multimodal Interaction (ICMI). pp. 173--180 (2025)

\bibitem{kipp2011assessing}
Kipp, M., Nguyen, Q., Heloir, A., Matthes, S.: Assessing the deaf user perspective on sign language avatars. In: Proceedings of the International ACM SIGACCESS Conference on Computers and Accessibility (ASSETS). pp. 107--114 (2011)

\bibitem{hanke-etal-2020-extending}
Konrad, R., Hanke, T., Langer, G., Blanck, D., Bleicken, J., Hofmann, I., Jeziorski, O., K{\"o}nig, L., K{\"o}nig, S., Nishio, R., Regen, A., Salden, U., Wagner, S., Worseck, S., B{\"o}se, O., Jahn, E., Schulder, M.: Meine dgs -- annotiert. {\"o}ffentliches korpus der deutschen geb{\"a}rdensprache, 3. release / my dgs -- annotated. public corpus of german sign language, 3rd release (2020). \doi{10.25592/dgs.corpus-3.0}, \url{https://doi.org/10.25592/dgs.corpus-3.0}, accessed: 2026-06-24

\bibitem{lee2019dancing}
Lee, H.Y., Yang, X., Liu, M.Y., Wang, T.C., Lu, Y.D., Yang, M.H., Kautz, J.: Dancing to music. Advances in Neural Information Processing systems (NeurIPS)  \textbf{32} (2019)

\bibitem{li2017learning}
Li, T., Bolkart, T., Black, M.J., Li, H., Romero, J.: Learning a model of facial shape and expression from 4d scans. In: Transactions on Graphics (TOG). vol.~36, pp. 194--1. ACM (2017)

\bibitem{li2025uni}
Li, Z., Zhou, W., Zhao, W., Wu, K., Hu, H., Li, H.: Uni-sign: Toward unified sign language understanding at scale. International Conference on Learning Representations (ICLR)  (2025)

\bibitem{DianeSignCompare}
Lillo-Martin, D.C., Gajewski, J.: One grammar or two? sign languages and the nature of human language. Wiley Interdisciplinary Reviews. Cognitive Science  \textbf{5}(4),  387–401 (2014)

\bibitem{lin2025handdiffuse}
Lin, P.: Handdiffuse: generative controllers for two-hand interactions via diffusion models. In: Proceedings of the AAAI Conference on Artificial Intelligence. vol.~39, pp. 5280--5288 (2025)

\bibitem{liu2023flow}
Liu, X., Gong, C., Liu, Q.: Flow straight and fast: Learning to generate and transfer data with rectified flow. In: International Conference on Learning Representations (ICLR) (2023)

\bibitem{teaser}
Liu, Y., Zhu, L., Lin, L., Zhu, Y., Zhang, A., Li, Y.: Teaser: Token enhanced spatial modeling for expressions reconstruction. In: International Conference on Learning Representations (ICLR) (2025)

\bibitem{low2025sage}
Low, J., Sincan, O.M., Bowden, R.: Sage: Segment-aware gloss-free encoding for token-efficient sign language translation. In: International Conference on Computer Vision Workshops (ICCVW). IEEE (2025)

\bibitem{he2025hands}
Low, J., Walsh, H., Sincan, O.M., Bowden, R.: Hands-on: Segmenting individual signs from continuous sequences. In: IEEE International Conference on Automatic Face and Gesture Recognition (FG) (2025)

\bibitem{moryossef2023linguistically}
Moryossef, A., Jiang, Z., M{\"u}ller, M., Ebling, S., Goldberg, Y.: Linguistically motivated sign language segmentation. In: Conference on Empirical Methods in Natural Language Processing (EMNLP) (2023)

\bibitem{moryossef2020real}
Moryossef, A., Tsochantaridis, I., Aharoni, R., Ebling, S., Narayanan, S.: Real-time sign language detection using human pose estimation. In: European Conference on Computer Vision Workshops (ECCVW). pp. 237--248. Springer (2020)

\bibitem{papineni2002bleu}
Papineni, K., Roukos, S., Ward, T., Zhu, W.J.: Bleu: a method for automatic evaluation of machine translation. In: Proceedings of the Annual Meeting of the Association for Computational Linguistics (ACL). pp. 311--318 (2002)

\bibitem{pavlakos2019expressive}
Pavlakos, G., Choutas, V., Ghorbani, N., Bolkart, T., Osman, A.A., Tzionas, D., Black, M.J.: Expressive body capture: 3d hands, face, and body from a single image. In: Proceedings of the IEEE/CVF Conference on Computer Vision and Pattern Recognition (CVPR). pp. 10975--10985 (2019)

\bibitem{pavlakos2024reconstructing}
Pavlakos, G., Shan, D., Radosavovic, I., Kanazawa, A., Fouhey, D., Malik, J.: Reconstructing hands in 3d with transformers. In: Proceedings of the IEEE/CVF Conference on Computer Vision and Pattern Recognition (CVPR). pp. 9826--9836 (2024)

\bibitem{Signvsspoken}
Perniss, P., Pfau, R., Steinbach, M.: Can't you see the difference? sources of variation in sign language structure. In: Visible variation: Cross-linguistic studies in sign language narratives, pp. 1--34. Mouton de Gruyter (2007)

\bibitem{petrovich2021action}
Petrovich, M., Black, M.J., Varol, G.: Action-conditioned 3d human motion synthesis with transformer vae. In: Proceedings of the IEEE/CVF International Conference on Computer Vision (ICCV). pp. 10985--10995 (2021)

\bibitem{petrovich2022temos}
Petrovich, M., Black, M.J., Varol, G.: Temos: Generating diverse human motions from textual descriptions. In: Proceedings of the European Conference on Computer Vision (ECCV). pp. 480--497. Springer (2022)

\bibitem{pfaunonmanuals}
Pfau, R., Quer, J.: Nonmanuals: their grammatical and prosodic roles. Sign Languages pp. 381--402 (2010)

\bibitem{plappert2016kit}
Plappert, M., Mandery, C., Asfour, T.: The kit motion-language dataset. Big data  \textbf{4}(4),  236--252 (2016)

\bibitem{potamias2025wilor}
Potamias, R.A., Zhang, J., Deng, J., Zafeiriou, S.: Wilor: End-to-end 3d hand localization and reconstruction in-the-wild. In: Proceedings of the IEEE/CVF Conference on Computer Vision and Pattern Recognition (CVPR). pp. 12242--12254 (2025)

\bibitem{punnakkal2021babel}
Punnakkal, A.R., Chandrasekaran, A., Athanasiou, N., Quiros-Ramirez, A., Black, M.J.: Babel: Bodies, action and behavior with english labels. In: Proceedings of the IEEE/CVF Conference on Computer Vision and Pattern Recognition (CVPR). pp. 722--731 (2021)

\bibitem{9413817}
Renz, K., Stache, N.C., Albanie, S., Varol, G.: Sign language segmentation with temporal convolutional networks. In: IEEE International Conference on Acoustics, Speech and Signal Processing (ICASSP). pp. 2135--2139 (2021)

\bibitem{renz2021sign}
Renz, K., Stache, N.C., Fox, N., Varol, G., Albanie, S.: Sign segmentation with changepoint-modulated pseudo-labelling. In: Proceedings of the IEEE/CVF Conference on Computer Vision and Pattern Recognition Workshops (CVPRW). pp. 3403--3412 (2021)

\bibitem{romero2017embodied}
Romero, J., Tzionas, D., Black, M.J.: Embodied hands: modeling and capturing hands and bodies together. In: Transactions on Graphics (TOG). vol.~36, pp. 1--17. ACM (2017)

\bibitem{5457527}
Santemiz, P., Aran, O., Saraclar, M., Akarun, L.: Automatic sign segmentation from continuous signing via multiple sequence alignment. In: International Conference on Computer Vision Workshops (ICCVW). pp. 2001--2008 (2009)

\bibitem{sarandi2024neural}
S{\'a}r{\'a}ndi, I., Pons-Moll, G.: Neural localizer fields for continuous 3d human pose and shape estimation. Advances in Neural Information Processing systems (NeurIPS)  \textbf{37},  140032--140065 (2024)

\bibitem{saunders2020adversarial}
Saunders, B., Camgoz, N.C., Bowden, R.: Adversarial training for multi-channel sign language production. In: British Machine Vision Conference (BMVC). British Machine Vision Association (2020)

\bibitem{saunders2020progressive}
Saunders, B., Camgoz, N.C., Bowden, R.: Progressive transformers for end-to-end sign language production. In: Proceedings of the European Conference on Computer Vision (ECCV). pp. 687--705. Springer (2020)

\bibitem{saunders2021continuous}
Saunders, B., Camgoz, N.C., Bowden, R.: Continuous 3d multi-channel sign language production via progressive transformers and mixture density networks. International Journal of Computer Vision (IJCV)  \textbf{129}(7),  2113--2135 (2021)

\bibitem{saunders2021mixed}
Saunders, B., Camgoz, N.C., Bowden, R.: Mixed signals: Sign language production via a mixture of motion primitives. In: Proceedings of the IEEE/CVF International Conference on Computer Vision (ICCV). pp. 1919--1929 (2021)

\bibitem{saunders2022signing}
Saunders, B., Camgoz, N.C., Bowden, R.: Signing at scale: Learning to co-articulate signs for large-scale photo-realistic sign language production. In: Proceedings of the IEEE/CVF Conference on Computer Vision and Pattern Recognition (CVPR). pp. 5141--5151 (2022)

\bibitem{shafir2024human}
Shafir, Y., Tevet, G., Kapon, R., Bermano, A.H.: Human motion diffusion as a generative prior. In: International Conference on Learning Representations (ICLR) (2024)

\bibitem{shlizerman2018audio}
Shlizerman, E., Dery, L., Schoen, H., Kemelmacher-Shlizerman, I.: Audio to body dynamics. In: Proceedings of the IEEE/CVF Conference on Computer Vision and Pattern Recognition (CVPR). pp. 7574--7583 (2018)

\bibitem{sincan2025gloss}
Sincan, O.M., Low, J.H., Asasi, S., Bowden, R.: Gloss-free sign language translation: An unbiased evaluation of progress in the field. Computer Vision and Image Understanding (CVIU) p. 104498 (2025)

\bibitem{stokoe2001language}
Stokoe, W.C.: Language in hand: Why sign came before speech. Gallaudet University Press (2001)

\bibitem{stoll2018sign}
Stoll, S., Camg{\"o}z, N.C., Hadfield, S., Bowden, R.: Sign language production using neural machine translation and generative adversarial networks. In: British Machine Vision Conference (BMVC). British Machine Vision Association (2018)

\bibitem{stoll2020text2sign}
Stoll, S., Camgoz, N.C., Hadfield, S., Bowden, R.: Text2sign: towards sign language production using neural machine translation and generative adversarial networks. International Journal of Computer Vision (IJCV)  \textbf{128}(4),  891--908 (2020)

\bibitem{stoll2022there}
Stoll, S., Mustafa, A., Guillemaut, J.Y.: There and back again: 3d sign language generation from text using back-translation. In: Proceedings of the International Conference on 3D Vision (3DV). pp. 187--196. IEEE (2022)

\bibitem{symeonidis2026m3t}
Symeonidis-Herzig, A., Low, J., Sincan, O.M., Bowden, R.: M3t: Discrete multi-modal motion tokens for sign language production. arXiv preprint arXiv:2603.23617  (2026)

\bibitem{tang2022gloss}
Tang, S., Hong, R., Guo, D., Wang, M.: Gloss semantic-enhanced network with online back-translation for sign language production. In: Proceedings of the ACM International Conference on Multimedia (MM). pp. 5630--5638 (2022)

\bibitem{tang2025gloss}
Tang, S., Xue, F., Wu, J., Wang, S., Hong, R.: Gloss-driven conditional diffusion models for sign language production. ACM Transactions on Multimedia Computing, Communications and Applications (TOMM)  \textbf{21}(4),  1--17 (2025)

\bibitem{tanzer2024youtube}
Tanzer, G., Zhang, B.: Youtube-sl-25: A large-scale, open-domain multilingual sign language parallel corpus. International Conference on Learning Representations (ICLR)  (2025)

\bibitem{tevet2022motionclip}
Tevet, G., Gordon, B., Hertz, A., Bermano, A.H., Cohen-Or, D.: Motionclip: Exposing human motion generation to clip space. In: Proceedings of the European Conference on Computer Vision (ECCV). pp. 358--374. Springer (2022)

\bibitem{tevethuman}
Tevet, G., Raab, S., Gordon, B., Shafir, Y., Cohen-or, D., Bermano, A.H.: Human motion diffusion model. In: International Conference on Learning Representations (ICLR) (2023)

\bibitem{varol2022scaling}
Varol, G., Momeni, L., Albanie, S., Afouras, T., Zisserman, A.: Scaling up sign spotting through sign language dictionaries. International Journal of Computer Vision (IJCV)  \textbf{130}(6),  1416--1439 (2022)

\bibitem{walshsign}
Walsh, H.T., Saunders, B., Bowden, R.: Sign stitching: A novel approach to sign language production. In: British Machine Vision Conference (BMVC). British Machine Vision Association (2024)

\bibitem{wang2025advanced}
Wang, C., Deng, Z., Jiang, Z., Yin, Y., Shen, F., Cheng, Z., Ge, S., Gan, S., Gu, Q.: Advanced sign language video generation with compressed and quantized multi-condition tokenization. Advances in Neural Information Processing systems (NeurIPS)  \textbf{38},  79519--79545 (2026)

\bibitem{wanggenerative}
Wang, X., Zhou, B., Curless, B., Kemelmacher-Shlizerman, I., Holynski, A., Seitz, S.: Generative inbetweening: Adapting image-to-video models for keyframe interpolation. In: International Conference on Learning Representations (ICLR) (2025)

\bibitem{wang2020learning}
Wang, Z., Yu, P., Zhao, Y., Zhang, R., Zhou, Y., Yuan, J., Chen, C.: Learning diverse stochastic human-action generators by learning smooth latent transitions. In: Proceedings of the AAAI Conference on Artificial Intelligence. vol.~34, pp. 12281--12288 (2020)

\bibitem{xing2024dynamicrafter}
Xing, J., Xia, M., Zhang, Y., Chen, H., Yu, W., Liu, H., Liu, G., Wang, X., Shan, Y., Wong, T.T.: Dynamicrafter: Animating open-domain images with video diffusion priors. In: Proceedings of the European Conference on Computer Vision (ECCV). pp. 399--417. Springer (2024)

\bibitem{yin2024t2s}
Yin, A., Li, H., Shen, K., Tang, S., Zhuang, Y.: T2s-gpt: Dynamic vector quantization for autoregressive sign language production from text. In: Proceedings of the Annual Meeting of the Association for Computational Linguistics (ACL). pp. 3345--3356 (2024)

\bibitem{yu2020structure}
Yu, P., Zhao, Y., Li, C., Yuan, J., Chen, C.: Structure-aware human-action generation. In: Proceedings of the European Conference on Computer Vision (ECCV). pp. 18--34. Springer (2020)

\bibitem{yu2024signavatars}
Yu, Z., Huang, S., Cheng, Y., Birdal, T.: Signavatars: A large-scale 3d sign language holistic motion dataset and benchmark. In: Proceedings of the European Conference on Computer Vision (ECCV). pp. 1--19. Springer (2024)

\bibitem{zelinka2020neural}
Zelinka, J., Kanis, J.: Neural sign language synthesis: Words are our glosses. In: Proceedings of the IEEE/CVF Winter Conference on Applications of Computer Vision (WACV). pp. 3395--3403 (2020)

\bibitem{zhang2025guava}
Zhang, D., Liu, Y., Lin, L., Zhu, Y., Li, Y., Qin, M., Li, Y., Wang, H.: Guava: Generalizable upper body 3d gaussian avatar. In: Proceedings of the IEEE/CVF International Conference on Computer Vision (ICCV). pp. 14205--14217 (2025)

\bibitem{zhang2023generating}
Zhang, J., Zhang, Y., Cun, X., Zhang, Y., Zhao, H., Lu, H., Shen, X., Shan, Y.: Generating human motion from textual descriptions with discrete representations. In: Proceedings of the IEEE/CVF Conference on Computer Vision and Pattern Recognition (CVPR). pp. 14730--14740 (2023)

\bibitem{zhang2024motiondiffuse}
Zhang, M., Cai, Z., Pan, L., Hong, F., Guo, X., Yang, L., Liu, Z.: Motiondiffuse: Text-driven human motion generation with diffusion model. IEEE Transactions on Pattern Analysis and Machine Intelligence (TPAMI)  \textbf{46}(6),  4115--4128 (2024)

\bibitem{zhou2023gloss}
Zhou, B., Chen, Z., Clap{\'e}s, A., Wan, J., Liang, Y., Escalera, S., Lei, Z., Zhang, D.: Gloss-free sign language translation: Improving from visual-language pretraining. In: Proceedings of the IEEE/CVF International Conference on Computer Vision (ICCV). pp. 20871--20881 (2023)

\bibitem{zhou2021improving}
Zhou, H., Zhou, W., Qi, W., Pu, J., Li, H.: Improving sign language translation with monolingual data by sign back-translation. In: Proceedings of the IEEE/CVF Conference on Computer Vision and Pattern Recognition (CVPR). pp. 1316--1325 (2021)

\bibitem{zhou2019continuity}
Zhou, Y., Barnes, C., Lu, J., Yang, J., Li, H.: On the continuity of rotation representations in neural networks. In: Proceedings of the IEEE/CVF Conference on Computer Vision and Pattern Recognition (CVPR). pp. 5745--5753 (2019)

\bibitem{zuo2024signs}
Zuo, R., Potamias, R.A., Ververas, E., Deng, J., Zafeiriou, S.: Signs as tokens: A retrieval-enhanced multilingual sign language generator. In: Proceedings of the IEEE/CVF International Conference on Computer Vision (ICCV). pp. 23806--23816 (2025)

\bibitem{zuo2024simple}
Zuo, R., Wei, F., Chen, Z., Mak, B., Yang, J., Tong, X.: A simple baseline for spoken language to sign language translation with 3d avatars. In: Proceedings of the European Conference on Computer Vision (ECCV). pp. 36--54. Springer (2024)

\bibitem{zwitserlood2004synthetic}
Zwitserlood, I., Verlinden, M., Ros, J., Van Der~Schoot, S., Netherlands, T.: Synthetic signing for the deaf: Esign. In: Proceedings of the Conference and Workshop on Assistive Technologies for Vision and Hearing Impairment (CVHI). vol.~1 (2004)

\end{thebibliography}
\end{document}